%% file: main.tex
\documentclass[10pt]{article} % For LaTeX2e
\usepackage[preprint]{tmlr}

\input{math_commands.tex}

\usepackage{hyperref}
\usepackage{mathtools} % amsmath with fixes and additions
\usepackage{booktabs} % commands to create good-looking tables
\usepackage{tikz} % nice language for creating drawings and diagrams
\usepackage{algorithm,algcompatible,amsmath}
\usetikzlibrary{calc,angles,positioning,intersections,quotes,decorations.markings, arrows.meta, shapes.geometric, backgrounds, fit}

\usepackage{enumitem}
\setlist[itemize]{noitemsep,nolistsep}
\usepackage{caption}
\usepackage{multirow}						% tabular cells spanning multiple rows
\usepackage{amsfonts}						% blackboard math symbols
\usepackage{graphicx}						% figures
\usepackage{duckuments}						% sample images
\usepackage{bbm}
\usepackage{stfloats}
\usepackage{algorithm,algcompatible,amsmath}
\usetikzlibrary{calc,angles,positioning,intersections,quotes,decorations.markings}
\usepackage{tkz-euclide}
\algnewcommand\INPUT{\item[\textbf{Input:}]}%
\algnewcommand\OUTPUT{\item[\textbf{Output:}]}%
\algnewcommand\FUNCTION{\item[\textbf{function}]}%
\algnewcommand\ENDFUNCTION{\item[\textbf{end function}]}%
\usepackage{setspace}
\usepackage{mathtools}

\title{Scalable Generative Modeling of Weighted Graphs}

\author{\name Richard Williams \email rlwilliams34@ucla.edu \\
        \addr Department of Biostatistics \\
        University of California, Los Angeles 
        \AND
        \name Eric Nalisnick \email nalisnick@jhu.edu \\
        \addr Department of Computer Science\\
        Johns Hopkins University
        \AND
        \name Andrew Holbrook \email aholbroo@ucla.edu \\
        \addr Department of Biostatistics \\
        University of California, Los Angeles}

\begin{document}

\maketitle

\begin{abstract}
  Weighted graphs are ubiquitous throughout biology, chemistry, and the social sciences,  motivating the development of generative models for abstract weighted graph data using deep neural networks. However, most current deep generative models are either designed for unweighted graphs and are not easily extended to weighted topologies or incorporate edge weights without consideration of a joint distribution with topology. Furthermore, learning a distribution over weighted graphs must account for complex nonlocal dependencies between both the edges of the graph and corresponding weights of each edge. We develop an autoregressive model BiGG-E, a nontrivial extension of the BiGG model, that learns a joint distribution over weighted graphs while still exploiting sparsity to generate a weighted graph with $n$ nodes and $m$ edges in $O((n + m)\log n)$ time. Simulation studies and experiments on a variety of benchmark datasets demonstrate that BiGG-E best captures distributions over weighted graphs while remaining scalable and computationally efficient.
  
\end{abstract}

%\doublespacing 

\section{Introduction}

Graphs are useful mathematical structures for representing data in a variety of domains, including biology, computer science, chemistry, and the social sciences, with applications ranging from modeling protein–protein interactions \citep{keretsu} to predicting time spent in traffic \citep{Stanojevic}.  A graph consists of a set of objects, called nodes, and their corresponding connections, called edges, which represent the graph's topology. Edges may contain additional information in the form of edge features, which can be categorical -- such as bond types in molecular graphs \citep{jo} -- or continuous -- such as branch lengths in phylogenetic trees \citep{semple}. Edge weights, in particular, are continuous single-dimensional edge features, and a graph with edge weights comprises a weighted graph. Weighted graphs find applications in many fields, such as neuroscience \citep{barjuan}, economics \citep{fagiolo}, social networks \citep{bellingeri}, and phylogenetics \citep{beast}. 

As weighted graphs prevail across domains, there is a need for generative models that capture both topological and edge weight distributions. More broadly, learning generative models over graph distributions is a vibrant area of research. Early approaches such as Erdős–Rényi \citep{erdos} and Barabási–Albert models \citep{albert} offer simple mechanisms but fail to capture the subtle dependencies between edges observed in real-world data. These limitations motivate the development of more expressive deep generative models capable of learning complex, nonlinear relationships. Despite recent advances, modeling graph distributions remains an ongoing challenge due to their combinatorial nature and the complex dependencies among edges. Furthermore, although incorporating edge weights appears straightforward, jointly modeling discrete topology and continuous weights introduces additional complexity, requiring the model to account for dependencies both within and between these two components.

Modern graph generative models -- including variational autoencoders (VAEs) \citep{kipf, grover}, graph neural networks \citep{grover}, autoregressive models \citep{you, liao, li, dai}, and score-based diffusion models \citep{niu, jo, vignac} -- primarily focus on unweighted graphs. Most limit their scope to modeling distributions over graph topology while offering limited insight into the joint modeling of topology and edge weights, and few provide scalable solutions for learning joint distributions over sparse weighted graphs. Furthermore, a significant computational bottleneck in graph generative modeling arises when jointly modeling all possible edge connections, which scales quadratically with the number of nodes. Models that attempt to jointly model all possible edge connections are computationally slow and infeasible for even moderately sized graphs. In contrast, autoregressive models factorize graph generation node-by-node using a sequential decision process. BiGG, ``Big Graph Generation'' \citep{dai}, augments this approach by directly generating the edge set of sparse graphs, scaling to graphs with tens of thousands of nodes.  However, existing autoregressive methods, including BiGG, remain limited to unweighted graphs.

To address the need for efficient generative modeling over large weighted graphs, we introduce BiGG-E (“BiGG-Extension”), an autoregressive model that jointly generates both graph topologies and edge weights while preserving the scalability of its unweighted predecessor, BiGG. We benchmark BiGG-E against three alternatives: (1) Adjacency-LSTM (Adj-LSTM), a fully expressive but computationally inefficient model parameterized with a  Long Short-Term Memory (LSTM; \citet{hochreiter}) cell; (2) BiGG-MLP, a naive extension that appends encodings of weights to BiGG using a multilayer perceptron (MLP, \citet{rumelhart}); and (3) BiGG+GCN, a two-stage model that decouples topology and weight generation.

Our contributions are as follows: 

\begin{itemize}
   \item We propose BiGG-E, an application-agnostic generative model that learns joint distributions over sparse weighted graphs.
    \item We empirically demonstrate that BiGG-E maintains the efficient scaling of BiGG while outperforming BiGG-MLP, Adj-LSTM, and BiGG+GCN.
    \item All BiGG extensions are orders of magnitude faster than Adj-LSTM and SparseDiff, a diffusion-model competitor.
   \item We directly evaluate the joint and marginal generative performance of all models on an array of weighted graph distributions.
\end{itemize}

\section{Background}

\subsection{Data} \label{data}  
Let $\mathcal{G} = \{G_1, \dots, G_{|\mathcal{G}|}\}$ be an independent sample of weighted graphs from an unknown data-generating distribution $p(G_i)$, for $i = 1, \dots, |\mathcal{G}|$. Each weighted graph is defined as $G_i = (V_i, E_i, W_i)$, where $V_i = \{v_1, \dots, v_{n_i}\}$ is the set of $|V_i| = n_i$ nodes, $E_i \subseteq V_i \times V_i$ is the set of $|E_i| = m_i $ edges, and $W_i: V_i \times V_i \to \mathbb{R}^+$ maps edges to positive edge weights. For notational simplicity, we drop the subscript $i$, under the assumption that the graphs $G_i \in \mathcal{G}$ are independent and identically distributed. 

For any edge $(v_i, v_j) \in E$, the edge weight is $W(v_i, v_j) = w_{ij}$; otherwise, it is zero. The weighted adjacency matrix $\mathbf{W} \in \mathbb{R}^{n \times n}$ has entries $W(v_i, v_j)$. In the unweighted case, edge weights are $1$ when $(v_i, v_j) \in E$ and $0$ otherwise. We denote the unweighted adjacency matrix by $\mathbf{A}$, which encodes the graph's topology, and use $\mathbf{W}$ to specifically refer to the weighted adjacency matrix. 

A weighted graph $G$ under node ordering $\pi$ is represented by its permuted weighted adjacency matrix $\mathbf{W}^{\pi}$, from which the probability of observing $G$ is given by $p(G) = p(|V| = n) \sum_{\pi} p(\mathbf{W}^{\pi(G)})$ \citep{dai}. Because summing over all $n!$ node permutations quickly becomes intractable, we follow \citet{liao} and assume a single canonical ordering $\pi$, yielding the lower bound estimate $p(G) \simeq p(|V| = n)p(\mathbf{W}^{\pi(G)})$. Following \citet{dai}, we estimate $p(|V| = n)$ using a multinomial distribution over node counts in the training set, and model $p(\mathbf{W}^{\pi(G)})$ with deep autoregressive neural networks parameterized by $\boldsymbol{\theta}$, denoted by $p_{\boldsymbol{\theta}}(\mathbf{W}^{\pi(G)})$.  We assume all graphs are under the canonical ordering $\pi(G)$ and omit this notation moving forward.

\subsection{Related Work}

\paragraph{Weighted Graph Generative Models} Although various models incorporate edge and node features in the graph generative process, these features are typically categorical \citep{kipf, li, kawai} or are tailored to a specific class of graphs, such as protein graphs \citep{ingraham}.  Furthermore, previous work on autoregressive models \citep{you, liao, dai} focuses exclusively on unweighted graphs. Most implementations on weighted graphs provide limited insight into the incorporation of edge weights. Graphite \citep{grover} proposes modeling weighted graphs by parameterizing a Gaussian random variable, which introduces the possibility of infeasible negative weights. Although score-based models incorporate edge features into the graph generative process, these features are typically categorical \citep{vignac} or rely on thresholding to produce a weighted adjacency matrix $\mathbf{W}$ and only evaluate performance on the binarized adjacency matrix \citep{niu}. 

\paragraph{Scalability} Scaling generative models to graphs with thousands of nodes is an ongoing challenge, as the adjacency matrix $\mathbf{A}$ has $\mathcal{O}(n^2)$ entries. In addition, many VAE \citep{grover} and diffusion \citep{niu, vignac} models utilize graph neural networks, which perform convolutions over the entire adjacency matrix of the graph. SparseDiff \citep{sparsediff} is a scalable diffusion model on sparse graphs, but only scales to graphs with hundreds of nodes, while we are interested in scaling to thousands of nodes.

Autoregressive models currently scale best with large graphs. While GraphRNN \citep{you} trains in $\mathcal{O}(n^2)$ time despite using a breadth-first search ordering scheme to reduce computational overhead, GRAN \citep{liao} trains in $\mathcal{O}(n)$ time by generating blocks of nodes of the graph at a time, but trades this gain in scalability for worsened sample quality as the model estimates edge densities per block of nodes. BiGG \citep{dai}  leverages the sparsity of many real-world graphs and directly generates the edge set $\{e_k\}$ of $\mathbf{A}$:

\begin{equation} \label{biggedge}
p_{\boldsymbol{\theta}}(\mathbf{A}) = \prod_{k = 1}^{m} p_{\boldsymbol{\theta}}(e_{k}|\{e_{l : l < k}\}).
\end{equation}

BiGG trains on the order $\mathcal{O}(\log n)$ time, generates an unweighted graph in $\mathcal{O}((n+m) \log n)$ time, and scales to graphs with up to 50K nodes.  Currently, BiGG and other autoregressive models remain limited to unweighted graphs, precluding the sampling of edge weights. These limitations motivate the need for a scalable autoregressive model capable of modeling joint distributions over weighted graphs.

\section{Methods and Contributions}

\subsection{Joint Modeling of Topology and Edge Weights} \label{sec:joint}

Previous autoregressive models produce unweighted graphs either by directly generating $\mathbf{A}$  \citep{you}, or by directly generating the edge set \citep{dai}. However, our models learn over \textit{weighted} adjacency matrices $\mathbf{W}$. As such, we first define a joint distribution over the existence of an edge $e$ and its corresponding edge weight $w$. To do so, note that as $w$ is only sampled when $e$ exists, we can naturally factor the joint probability $p_{\boldsymbol{\theta}}(e, w)$ of observing a weighted edge $(e, w)$ as 

\begin{equation} \label{jfactor}
p_{\boldsymbol{\theta}}(e, w) = p_{\boldsymbol{\theta}}(e)p_{\boldsymbol{\theta}}(w|e),
\end{equation}

where $p_{\boldsymbol{\theta}}(e)$ is the parameterized Bernoulli probability that an edge exists between two nodes, and $p_{\boldsymbol{\theta}}(w|e)$ is the probability of drawing a corresponding weight given that $e$ exists. Since $w$ is assumed to be continuous, let $p_{\boldsymbol{\theta}}(w|e)$ represent the distribution of the random variable $w$ with parameterized density $f_{\boldsymbol{\theta}}(w)$.  In the case where no edge exists and $e = 0$, set $w = 0$ with probability 1; otherwise, if an edge exists and $e = 1$, draw a corresponding weight from a conditional distribution $p_{\boldsymbol{\theta}}(w|e)$. 

We parameterize the conditional distribution $p_{\boldsymbol{\theta}}(w | e)$ as  a normal random variable $\epsilon | e \sim N(\mu, \sigma)$ transformed with the softplus function $\textrm{Softplus}(\epsilon) = \log (1 + \exp(\epsilon))$. In our experience, such a transformation of a normal random variable performs best with gradient-based optimization by providing enough flexibility in modeling distributions, where work such as \citet{rodriguez} empirically demonstrates that a probit transformation of a random normal variable provides a prior capable of generating a rich class of distributions. To ensure positivity of the weights, the softplus function maps each value from the normal distribution to a positive real number. Other candidate distributions, such as the gamma and log-normal distributions, are more challenging to implement because of the complexity of the likelihood in the former and the heavy right-tailedness in the latter. Thus, with the softplus-normal conditional density placed on the weights, the term $p_{\boldsymbol{\theta}}(w|e)$ in Equation~\ref{jfactor} is equal to

\begin{equation} \label{eq:spnorm}
    p_{\boldsymbol{\theta}}(w | e) \propto \frac{1}{2\sigma^2} \exp\bigg[-\frac{1}{2\sigma^2}\big(\log(e^{w} - 1) - \mu\big)^{2}\bigg]
\end{equation}

up to a constant factor, where $\mu$ and $\sigma^2$ are functions of neural network parameters $\boldsymbol{\theta}$.

\subsection{Likelihood of a Weighted Adjacency Matrix}

There are two ways to parameterize the distribution $p_{\boldsymbol{\theta}}(\mathbf{W})$ over weighted graphs: first, we may consider the probability over all entries of $\mathbf{W}$ in a row-wise manner as

\begin{equation} \label{adjwprob}
p_{\boldsymbol{\theta}}(\mathbf{W}) = \prod_{i=1}^{n} \prod_{j=1}^{i-1} p_{\boldsymbol{\theta}}(W_{ij} | \{W_{kl}\}) = 
\prod_{i=1}^{n} \prod_{j=1}^{i-1}(1-p_{ij})^{1 - e_{ij}} \big[p_{ij} p_{\boldsymbol{\theta}}(w_{ij}|e_{ij})\big]^{e_{ij}},
\end{equation}

where $W_{ij}$ is the $(i,j)$-th entry of $\mathbf{W}$, $p_{ij} \equiv p_{ij}(\boldsymbol{\theta})$ is the estimated probability of an edge existing between nodes $v_i$ and $v_j$, $e_{ij} = 1$ when $(v_i, v_j) \in E$ and is otherwise 0, and $w_{ij} = W(v_i, v_j)$ is the weight of edge $e_{ij}$ whenever $(v_i, v_j) \in E$. Note each entry $W_{ij}$ is conditioned on all prior entries, denoted as  $\{W_{kl}\}$.

Next, similarly to how BiGG factors $p_{\boldsymbol{\theta}}(\mathbf{A})$ in Equation~\ref{biggedge}, we factor the weighted edge set of $\mathbf{W}$ as

\begin{equation} \label{biggwprob}
p_{\boldsymbol{\theta}}(\mathbf{W}) = \prod_{k = 1}^{m} p_{\boldsymbol{\theta}}(e_{k}|\{(e_{l}, w_{l})_{: l < k}\}) \cdot p_{\boldsymbol{\theta}}(w_{k}|e_k, \{(e_{l}, w_{l})_{: l < k}\}),
\end{equation}

noting all edges up to and including $e_k$ condition the generation of weight $w_k$. We substitute Equation~\ref{eq:spnorm} into the terms $p_{\boldsymbol{\theta}}(w_{ij}|e_{ij})$ and $p_{\boldsymbol{\theta}}(w_{k}|e_k, \{(e_{l}, w_{l})_{: l < k}\})$ of Equations~\ref{adjwprob} and \ref{biggwprob}, respectively, and maximize the log-likelihood $\mathcal{L}(\boldsymbol{\theta}; \mathbf{W})$ to train our models on weighted graphs. More details of the derivation of Equations~\ref{adjwprob} and \ref{biggwprob} and the objective function are given in Appendix \ref{sec:probcalc}. 

\subsection{Models}

Our main contributions use autoregressive models, which are well-suited for graph generation as they explicitly capture dependencies among edges and, in our case, their weights. BiGG-E extends the original BiGG model by maintaining two states during generation: a topological state inherited from BiGG and a weight state that encodes all previously generated edge weights. By leveraging both states, BiGG-E jointly and autoregressively predicts the weighted edge set of $\mathbf{W}$. We begin by reviewing the BiGG architecture before detailing our extensions in BiGG-E. Additional architectural details and an expanded BiGG review are provided in Appendices~\ref{sec:architect} and \ref{sec:biggexpand}.

%\subsubsection{Review of BiGG \citep{dai}} \label{sec:bigg}
\subsubsection{\texorpdfstring{Review of BiGG \citep{dai}}{Review of BiGG}} \label{sec:bigg}

BiGG generates an unweighted graph with an algorithm consisting of two main components, both of which train in $\mathcal{O}(\log n)$ time: (1) row generation, where BiGG generates each row of the lower half of $\mathbf{A}$ using a binary decision tree; and (2) row conditioning, where BiGG deploys a hierarchical data maintenance structure called a Fenwick tree \citep{fenwick} to condition the subsequent row generation on all previous rows.  

\paragraph{Row Generation} \label{sec:rowgen}

To sample edge connections for each node $v_u \in G$, BiGG adapts a procedure from R-MAT \citep{chakrabarti} to construct a binary decision tree $\mathcal{T}_{u}$, which identifies all edge connections with $v_u$ by recursively partitioning the candidate edge interval $[v_1,v_{u-1}]$ into halves.  Each node $t \in \mathcal{T}_{u}$ corresponds to a subinterval $[v_i, v_k]$ of length $l_t = k - i + 1$, which is split into left and right halves: $\text{lch}(t) = [v_i,\ v_{i + \lfloor l_t / 2 \rfloor}]$ and $\text{rch}(t) = [v_{i + \lfloor l_t / 2 \rfloor + 1},\ v_k]$.

If an edge exists in lch($t$), the model recurses into that interval until reaching a singleton interval $[v_j, v_j]$, which represents an edge connection with $v_j$.   After completing the left subtree, the model recurses into rch($t$), conditioned on all edge connections -- if any -- formed in lch($t$). BiGG conditions subsequent predictions on all prior interval splits using two context vectors: (1) predictions for $\textrm{lch}(t)$ use a top-down hidden state $\mathbf{h}^{\text{top}}_u(t)$, which sequentially encodes all left and right edge existence decisions made in $\mathcal{T}_{u}$ thus far; and (2) predictions for $\textrm{rch}(t)$ use a conditioned top-down hidden state $\mathbf{\hat{h}}^{top}_u(t)$ computed by merging $\mathbf{h}^{\text{top}}_u(t)$ with a bottom-up summary state of the generated left subtree, $\mathbf{h}^{\text{bot}}_u(\text{lch}(t))$. This merge is performed using a Tree-LSTM Cell \citep{tai}, which encodes relevant information from the top-down and bottom-up hidden states: $\mathbf{\hat{h}}^{top}_u(t) = \textrm{TreeCell}_{\boldsymbol{\theta}}(\mathbf{h}^{\text{top}}_u(t),  \mathbf{h}^{\text{bot}}_u(\textrm{lch}(t)))$.

Hence, constructing $\mathcal{T}_u$ is fully autoregressive with probability

\begin{equation} \label{eq:tuprobbigg}
p_{\boldsymbol{\theta}}(\mathcal{T}_u ) =  \prod_{t\in \mathcal{T}_u}p_{\boldsymbol{\theta}}(\textrm{lch}(t) | \mathbf{h}^{\text{top}}_u(t)) \cdot  p_{\boldsymbol{\theta}}(\textrm{rch(t)} | \mathbf{\hat{h}}^{\text{top}}_u(t)).
\end{equation}

\begin{figure}[t]
\centering
\begin{tikzpicture}[node distance=0.7cm and 2cm, every node/.style={font=\small}, edge/.style={thick}, dashededge/.style={thick, dashed, red}, level distance=0.5cm, sibling distance=0.75cm,scale=0.70]
\node[] (paramsk) at (0.3,-0.5) {};
\node[] (paramsk1) at (4.7,-0.5) {};

% Root node
\node (t0) at (0, 5) {$[v_1, v_8]$};

% Tree nodes
\node (t1) at ($(t0) + (-2.5, -2.5)$) {$[v_1, v_4]$};
\node (t2) at ($(t0) + (2.5,-2.5)$) {};

\node (t3) at ($(t1) + (-2.25, -3)$) {$[v_1, v_2]$};
\node (t4) at ($(t1) + (2.25, -3)$) {$[v_3, v_4]$};

\node (t7) at ($(t4) + (-2, -3)$) {};
\node (t8) at ($(t4) + (2, -3)$) {$v_4$};

\node (t9) at ($(t3) + (-2, -3)$) {$v_1$};
\node (t10) at  ($(t3) + (2, -3)$) {};

% Edges
\draw[->, thick, purple!50!white] (t0) -- (t1);
\draw[-, thick, dashed, orange!70!white] (t0) -- (t2);
\draw[->, thick, purple!50!white] (t1) -- (t3);
\draw[->, thick, orange!70!white] (t1) -- (t4);
\draw[-, thick, dashed, purple!50!white] (t4) -- (t7);
\draw[->, thick, orange!70!white] (t4) -- (t8);
\draw[->, thick, purple!50!white] (t3) -- (t9);
\draw[-, thick, dashed, orange!70!white] (t3) -- (t10);

\node[circle, draw, solid, fill=yellow, minimum size=0.4cm, inner sep=1.5pt] (l1) at ($(t1)+(0.3,-0.5)$) {};

\node (pointer) at  ($(t3)+(0.67,0.17)$) {};
\node (pointer2) at ($(l1)+(-0.80, 0.11)$) {};

\draw[->, thick, blue!60] (pointer) -- (l1);
\draw[->, thick, purple!50!white] (pointer2) -- (l1);

\node (phantomRightTop) at ($(t3)+(1.85,-0.18)$) {};
\node (phantomLeftBot) at ($(t9)+(-0.0,-0.0)$) {};

\begin{pgfonlayer}{background}
  \node[draw=blue!60, thick, rounded corners, fit=(phantomRightTop)(phantomLeftBot), inner sep=0.25cm] {};
\end{pgfonlayer}

\node (leg1) at ($(t1) + (7, -2.65)$) {$\mathbf{h_{u}^{top}}$};
\node (leg2) at ($(leg1) + (0.00, -0.75)$) {$\mathbf{\hat{h}_{u}^{top}}$};
\node (leg3) at ($(leg2) + (0.00, -0.75)$) {$\mathbf{h_{u}^{bot}}$};
\node (leg4) at ($(leg3) + (1.1, -0.75)$) {Tree-LSTM Cell};
\node (leg5) at ($(leg4) + (0.65, -0.75)$) {Left Subtree Summary};

\draw[->, thick, purple!50!white] ($(leg1) + (-1.2, 0)$) -- ($(leg1) + (-0.6, 0)$);
\draw[->, thick, orange!70!white] ($(leg2) + (-1.2, 0)$) -- ($(leg2) + (-0.6, 0)$);
\draw[->, thick, blue!60] ($(leg3) + (-1.2, 0)$) -- ($(leg3) + (-0.6, 0)$);

\node[circle, draw, solid, fill=yellow, minimum size=0.3cm, inner sep=1.5pt] (l2) at ($(leg4) + (-2, 0)$) {};

\node[rectangle, draw, color=blue!60] (l3) at ($(leg5) + (-2.65, 0)$) {};

\node (phantomRightTop2) at ($(leg1)+(-1,0.0)$) {};
\node (phantomLeftBot2) at ($(leg5)+(2.00,-0.0)$) {};

\begin{pgfonlayer}{background}
  \node[draw=gray, thick, rounded corners, dashed, fit=(phantomRightTop2)(phantomLeftBot2), inner sep=0.25cm] {};
\end{pgfonlayer}

\end{tikzpicture}
\captionof{figure}[Example of constructing $\mathcal{T}_{u}$ in the original BiGG model.]
{Example of constructing $\mathcal{T}_{u}$ in the original BiGG model. The interval $[v_1, v_8]$ is recursively partitioned via left-right decisions until reaching individual nodes (e.g., $v_1$ and $v_4$).  Dashed lines indicate no edge. Purple arrows show the top-down context vector $\mathbf{h}_u^{\text{top}}$ used for left-child edge predictions; orange arrows show the conditioned top-down context $\mathbf{\hat{h}}_u^{\text{top}}$ used for right-child predictions. At node $[v_1, v_4]$, a TreeLSTM merges the left sub-tree summary (blue) with $\mathbf{h}_u^{\text{top}}$ to produce $\mathbf{\hat{h}}_u^{\text{top}}$ for left subtree conditioning.}
\label{fig:topfig}
\end{figure}
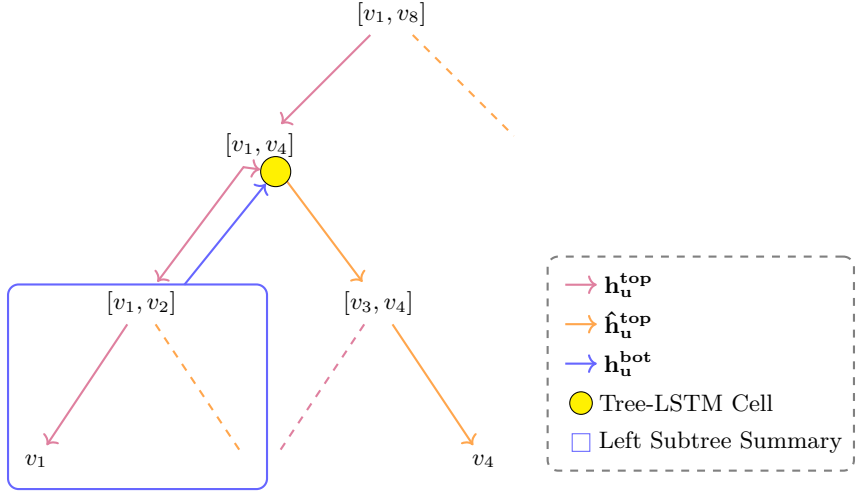

Figure~\ref{fig:topfig} illustrates an example of constructing $\mathcal{T}_u$ and visualizes use of the top-down and bottom-up states in predicting $\textrm{lch}(t)$ and $\textrm{rch}(t)$. Finally, note that as $\mathbf{h}_{u}^{\text{bot}}(t)$ is the bottom-up summary state summarizing the subtree rooted at node $t$, the bottom-up summary state at the root node $t_0$ summarizes the entire tree $\mathcal{T}_u$.

\paragraph{Fenwick Tree} \label{sec:fentapp} 

To condition each decision tree $\mathcal{T}_{u}$ on all prior trees $\mathcal{T}_{1}$  to $\mathcal{T}_{u-1}$, BiGG adopts the Fenwick tree \citep{fenwick} to efficiently summarize all previously generated rows in $\mathbf{A}$. The Fenwick tree at row $u$ has $\lfloor \log(u - 1) \rfloor + 1$ levels, where the base level (leaves)  of the tree are states $\mathbf{h}_{u}^{bot}(t_0)$ summarizing the edge connections formed in each $\mathcal{T}_{j}$, $j = 1, \dots, u-1$. Higher levels of the Fenwick tree merge these states to produce aggregated summaries across multiple rows. Letting $\mathbf{g}^{i}_j$ denote the state at the $j$-th node on the $i$-th level, each non-leaf node of the Fenwick tree merges its two children using a Tree-LSTM cell as

\begin{equation} \label{eq:fentree}
\mathbf{g}_{j}^{i} = \textrm{TreeCell}_{\boldsymbol{\theta}}^{\text{row}}(\mathbf{g}^{i-1}_{2j-1}, \mathbf{g}^{i-1}_{2j}),
\end{equation}

where $1 \le i \le \lfloor \log(u - 1) \rfloor + 1$, $1 \le j \le \lfloor \frac{u}{2^i} \rfloor$, and $\mathbf{g}_{j}^{0}$ is the bottom-up summary state of $\mathcal{T}_j$. Finally, to summarize all rows $1$ to $u -1$, the model iteratively applies a Tree-LSTM cell to produce a row summary hidden state $\mathbf{h}^{\text{row}}_u$:

\begin{equation} \label{eq:fenrow}
\mathbf{h}^{\text{row}}_u = \textrm{TreeCell}_{\boldsymbol{\theta}}^{\textrm{summary}}\bigg(\bigg[\mathbf{g}^{i}_{\floor{\frac{u}{2^{i}}}}\ \textrm{where}\ u\ \&\ 2^{i} = 2^{i} \bigg]\bigg),
\end{equation}

where \& is the bit-level `and' operator, each $\mathbf{g}^{i}_{\floor{\frac{u}{2^{i}}}}$ encodes summaries of different groups of rows of $\mathbf{A}$, and $\mathbf{h}^{\text{row}}_u $ initializes $h_{u}^{\text{top}}(t_0)$ to condition construction of $\mathcal{T}_u$ so that $p_{\boldsymbol{\theta}}(\mathcal{T}_u) \equiv p_{\boldsymbol{\theta}}(\mathcal{T}_u|\mathcal{T}_1,\dots,\mathcal{T}_{u-1})$.

\paragraph{BiGG Training and Sampling Times} \label{sec:bigg-train}

The training procedure for BiGG consists of four steps, each running in $\mathcal{O}(\log n)$ time by parallelizing computations across rows. First, since trees $\mathcal{T}_u$ are summarized independently, their root-level summaries are computed in parallel by traversing each tree level by level from the leaves to the root. Second, the Fenwick tree is constructed from these root summaries in the same level-wise manner. Third, the model computes all row summaries $\mathbf{h}_{u}^{\text{row}}$ using the Fenwick tree. Finally, for each $\mathcal{T}_{u}$, the model computes all left and right edge interval existence probabilities level-by-level. 

For graph generation, the Fenwick tree requires $\mathcal{O}(n \log n)$ time to construct, since updates now occur sequentially across row. Provided the graph is sparse, i.e., $m = \mathcal{O}(n)$, the construction of all trees $\mathcal{T}_u$ requires $\mathcal{O}(m \log n)$ time. Thus, the total sampling time of a sparse unweighted graph is $\mathcal{O}((n + m )\log n)$.

\subsection{BiGG-E} \label{sec:bigge-desc}

BiGG-E incorporates a weight state used in tandem with the original topology state of the BiGG model to jointly predict weighted edges. We first describe the weight state, followed by the joint prediction framework.

\begin{algorithm}[t] 
    \caption{BiGG-E Weight Sampling and Embedding}
    \label{alg1}
  \begin{algorithmic}[1]
    \FUNCTION{embed\_weight($w_{k}$, $\mathbf{\mathbf{h}}_{k-1}^{\text{wt}}$)}
    \STATE $\mathbf{w}^{0}_{k} = \textrm{LSTM}_{\boldsymbol{\theta}}(w_{k})$
    \STATE Add $\mathbf{w}^{0}_{k}$ to Fenwick weight tree and update tree using Equation~\ref{eq:fentree}.
    \STATE $\mathbf{h}_{k}^{\text{wt}} = \textrm{TreeCell}_{\boldsymbol{\theta}}^{\text{summary}}\bigg(\bigg[\mathbf{w}^{i}_{\floor{\frac{k}{2^{i}}}}$ where $k$ \& $2^{i} = 2^{i} \bigg]\bigg)$
    \STATE Return $\mathbf{h}_{k}^{\text{wt}}$
    \ENDFUNCTION
    \Statex
    \FUNCTION{sample\_weight$(u, t, \mathbf{h}_{u}^{\text{top}}(t), \mathbf{h}_{k}^{\text{wt}})$}
      \STATE $\mathbf{h}^{\text{sum}}_{u, k}(t) = \textrm{TreeCell}_{\boldsymbol{\theta}}^{\textrm{merge}}(\mathbf{h}_{u}^{\text{top}}(t),  \mathbf{h}_{k}^{\text{wt}})$
      \STATE Set $\mu_{k+1} = f_\mu(\mathbf{h}^{\text{sum}}_{u, k}(t))$ and $\log \sigma^2_{k+1} = f_{\sigma^2}(\mathbf{h}^{\text{sum}}_{u, k}(t))$
      \STATE Sample $w_{k+1}$ from Section~\ref{sec:joint} using $\mu_{k+1}, \sigma^{2}_{k+1}$.
      \STATE $\mathbf{h}_{k+1}^{\text{wt}}$ = embed\_weight($w_{k+1}$, $\mathbf{h}_{k}^{\text{wt}})$
      \STATE Return $\vec{1}$, \{edge index $t$ represents\}, $w_{k+1}$, $\mathbf{h}_{k+1}^{\text{wt}}$
    \ENDFUNCTION
  \end{algorithmic}
\end{algorithm}

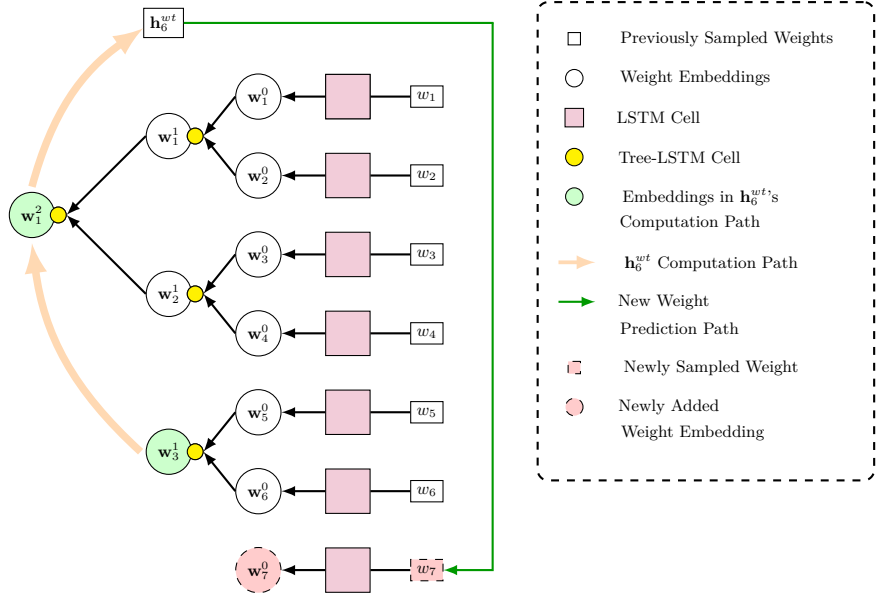
\begin{figure} [t]
\begin{center}
\begin{tikzpicture}[scale=0.7, transform shape, node distance=1.0cm and 1.0cm, every node/.style={font=\small}, highlight/.style={fill=yellow!40}, >=latex]

% Add TreeLSTM Nodes
\node[circle, draw, solid, fill=yellow, minimum size=0.3cm, inner sep=1pt] (t1) at (3.1, -2.15) {};
\node[circle, draw, solid, fill=yellow, minimum size=0.3cm, inner sep=1pt] (t2) at (3.1, -5.15) {};
\node[circle, draw, solid, fill=yellow, minimum size=0.3cm, inner sep=1pt] (t3) at (3.1, -8.15) {};
\node[circle, draw, solid, fill=yellow, minimum size=0.3cm, inner sep=1pt] (t4) at (0.5, -3.65) {};

% Raw weights
\node[rectangle, draw] (rw1) at (7.5,-1.4) {$w_1$};
\node[rectangle, draw] (rw2) at (7.5,-2.9) {$w_2$};
\node[rectangle, draw] (rw3) at (7.5,-4.4) {$w_3$};
\node[rectangle, draw] (rw4) at (7.5,-5.9) {$w_4$};
\node[rectangle, draw] (rw5) at (7.5,-7.4) {$w_5$};
\node[rectangle, draw] (rw6) at (7.5,-8.9) {$w_6$};
\node[rectangle, draw, fill=red!20!white, dashed] (rw7) at (7.5,-10.4) {$w_7$};

% Level 0 weight embeddings
\node[circle,draw] (w10) at (4.3,-1.4) {$\mathbf{w}_1^{0}$};
\node[circle,draw] (w20) at (4.3,-2.9) {$\mathbf{w}_2^{0}$};
\node[circle,draw] (w30) at (4.3,-4.4) {$\mathbf{w}_3^{0}$};
\node[circle,draw] (w40) at (4.3,-5.9) {$\mathbf{w}_4^{0}$};
\node[circle,draw] (w50) at (4.3,-7.4) {$\mathbf{w}_5^{0}$};
\node[circle,draw] (w60) at (4.3,-8.9) {$\mathbf{w}_6^{0}$};
\node[circle,draw,dashed,fill=red!20!white] (w70) at (4.3,-10.4) {$\mathbf{w}_7^{0}$};

% LSTM Block
\node[rectangle,draw,fill=purple!20!white, inner sep = 12pt] (lstm) at (6, -1.4) {};
\node[rectangle,draw,fill=purple!20!white, inner sep = 12pt] (lstm1) at (6, -1.4) {};
\node[rectangle,draw,fill=purple!20!white, inner sep = 12pt] (lstm2) at (6, -2.9) {};
\node[rectangle,draw,fill=purple!20!white, inner sep = 12pt] (lstm3) at (6, -4.4) {};
\node[rectangle,draw,fill=purple!20!white, inner sep = 12pt] (lstm4) at (6, -5.9) {};
\node[rectangle,draw,fill=purple!20!white, inner sep = 12pt] (lstm5) at (6, -7.4) {};
\node[rectangle,draw,fill=purple!20!white, inner sep = 12pt] (lstm6) at (6, -8.9) {};
\node[rectangle,draw,fill=purple!20!white, inner sep = 12pt] (lstm7) at (6, -10.4) {};

% Arrows Pointing to LSTM Embedings
\draw[thick] (rw1.west) -- (lstm1.east);
\draw[thick] (rw2.west) -- (lstm2.east);
\draw[thick] (rw3.west) -- (lstm3.east);
\draw[thick] (rw4.west) -- (lstm4.east);
\draw[thick] (rw5.west) -- (lstm5.east);
\draw[thick] (rw6.west) -- (lstm6.east);
\draw[thick] (rw7.west) -- (lstm7.east);

% Arrows Point from LSTM to Level 0:
\draw[thick, ->] (lstm1.west) -- (w10.east);
\draw[thick, ->] (lstm2.west) -- (w20.east);
\draw[thick, ->] (lstm3.west) -- (w30.east);
\draw[thick, ->] (lstm4.west) -- (w40.east);
\draw[thick, ->] (lstm5.west) -- (w50.east);
\draw[thick, ->] (lstm6.west) -- (w60.east);
\draw[thick, ->] (lstm7.west) -- (w70.east);

% Level 1 weight embeddings
\node[circle,draw] (w11) at (2.6,-2.15) {$\mathbf{w}_1^{1}$};
\node[circle,draw] (w21) at (2.6,-5.15) {$\mathbf{w}_2^{1}$};
\node[circle,draw,fill=green!20!white] (w31) at (2.6,-8.15) {$\mathbf{w}_3^{1}$};

% Level 0 to 1 Arrows
\draw[thick, ->] (w10.west) -- (t1.east);
\draw[thick, ->] (w20.west) -- (t1.east);
\draw[thick, ->] (w30.west) -- (t2.east);
\draw[thick, ->] (w40.west) -- (t2.east);
\draw[thick, ->] (w50.west) -- (t3.east);
\draw[thick, ->] (w60.west) -- (t3.east);

% Level 2 weight embeddings
\node[circle,draw,fill=green!20!white] (w12) at (0,-3.65) {$\mathbf{w}_1^{2}$};

% Add TreeLSTM Nodes
\node[circle, draw, solid, fill=yellow, minimum size=0.3cm, inner sep=1pt] (t1) at (3.1, -2.15) {};
\node[circle, draw, solid, fill=yellow, minimum size=0.3cm, inner sep=1pt] (t2) at (3.1, -5.15) {};
\node[circle, draw, solid, fill=yellow, minimum size=0.3cm, inner sep=1pt] (t3) at (3.1, -8.15) {};
\node[circle, draw, solid, fill=yellow, minimum size=0.3cm, inner sep=1pt] (t4) at (0.5, -3.65) {};

% Arrows level 1 to level 2
\draw[thick, ->] (w11.west) -- (t4.east);
\draw[thick, ->] (w21.west) -- (t4.east);

% hwt node
\node[rectangle, draw] (hwt) at (2.5, 0.0) {$\mathbf{h}_{6}^{wt}$};

% Arrows computation pathway
\draw[thick, ->, color=orange!30!white, line width = 1mm] ($(w31.west) + (-0.1, 0)$) to[bend left=20] ($(w12.south) + (0, -0.1)$);

\draw[thick, ->, color=orange!30!white, line width = 1mm] ($(w12.north) + (0, 0.1)$) to[bend left=20] ($(hwt.west) + (0, -0.1)$);

% Path Arrow to Predict Next weight
\node[] (turn) at (8.75, 0.0) {};
\draw[thick, color=green!60!black] (hwt.east) -- ($(turn) + (0.0145, 0)$);
\draw[thick, color=green!60!black, ->] ($(turn.south) + (0, 0.135)$) |- (rw7.east);

% Legend
\begin{scope}[xshift=10.3cm, yshift=-0.3cm]

\node[rectangle,draw] (A) at (0, 0.0) {};
\node[] (desc2) at (2.9, 0.0) {Previously Sampled Weights};

\node[circle,draw] (B) at (0, -0.75) {};
\node[] (desc1) at (2.3, -0.75) {Weight Embeddings};

\node[rectangle,draw,fill=purple!20!white, inner sep = 5pt] (C) at (0, -1.5) {};
\node[] (desc5) at (1.6, -1.5) {LSTM Cell};

\node[circle,draw,fill=yellow] (D) at (0, -2.25) {};
\node[] (desc3) at (2, -2.25) {Tree-LSTM Cell};

\node[circle,draw,fill=green!20!white] (E) at (0, -3) {};
\node[] (desc4) at (2.4, -3) {Embeddings in $\mathbf{h}_6^{wt}$'s};
\node[] (desc4b) at (2.2, -3.5) {Computation Path};

\draw[thick, ->, color=orange!30!white, line width = 0.5mm] (-0.3, -4.25) -- (0.4, -4.25);
\node[] (desc5) at (2.6, -4.25) {$\mathbf{h}_6^{wt}$ Computation Path};

\draw[thick, ->, color=green!60!black] (-0.3, -5.0) -- (0.4, -5.0);
\node[] (desc6) at (1.7, -5.0) {New Weight};
\node[] (desc6b) at (2, -5.5) {Prediction Path};

\node[rectangle,draw,dashed,fill=red!20!white] (A) at (0, -6.25) {};
\node[] (desc7) at (2.6, -6.25) {Newly Sampled Weight};

\node[circle,draw,dashed,fill=red!20!white] (B) at (0, -7) {};
\node[] (desc8) at (1.8, -7) {Newly Added};
\node[] (desc8b) at (2.25, -7.5) {Weight Embedding};

\node[] (phantomTL) at (0.0, 0.0) {};
\node[] (phantomBR) at (5.0, -7.7) {};

\begin{pgfonlayer}{background}
  \node[draw=black, thick, rounded corners, dashed, fit=(phantomTL)(phantomBR), inner sep=0.4cm] {};
\end{pgfonlayer}
\end{scope}

\end{tikzpicture}
\end{center}
\captionof{figure}[Fenwick weight tree state construction]
{Illustration of Fenwick weight tree state construction. Sampled weights $w_j$ are passed through an LSTM to obtain initial embeddings $\mathbf{w}_j^0$, which are recursively merged using Tree-LSTM cells (yellow nodes) to form higher-level summaries. To compute the current weight hidden state $\mathbf{h}_6^{wt}$, summaries for $w_1$ to $w_4$ ($\mathbf{w}_{1}^{2}$) and $w_5$ to $w_6$ ($\mathbf{w}_{1}^{2})$ are merged. This state is used to predict $w_7$, which is added to the Fenwick weight tree and updated accordingly.}
\label{fig:fenwtfig}
\end{figure}

\subsubsection{Edge Weight Prediction} \label{sec:bigge}

\paragraph{Constructing the Weight State} 

To preserve BiGG's training and sampling speed-ups, BiGG-E introduces a second Fenwick data structure -- referred to as the Fenwick \textit{weight} tree -- that summarizes edge weights during generation. This new tree mirrors the structure of the original Fenwick \textit{topology} tree used in BiGG, but is maintained separately to construct edge weight embeddings autoregressively. The Fenwick weight tree is similarly organized into $\floor{\log(k - 1)} + 1$ levels, where $k$ is the current number of weights in the graph and the 0-th level corresponds to initialized weight embeddings, computed using a single forward pass of an LSTM: $\mathbf{w}_k^{0} = \textrm{LSTM}_{\boldsymbol{\theta}}(w_k)$. Higher-level embeddings $\mathbf{w}_{j}^{i}$ in the Fenwick weight tree are computed using Equation~\ref{eq:fentree}, where $\mathbf{w}_{j}^{0}$ is now the initial embedding of the $j$-th weight in the graph. 

To obtain a summary state of all prior edge weights, we use Equation~\ref{eq:fenrow} to compute the summary weight state $\mathbf{h}^{wt}_{k}$ for weights $w_1$ to $w_k$ in $\mathcal{O}(\log k)$ steps. Algorithm~\ref{alg1} outlines the edge weight embedding procedure in the Fenwick weight tree. Figure~\ref{fig:fenwtfig} illustrates an example of the Fenwick weight tree embedding process, which updates the weight state recursively based on the weights generated so far.

%\begin{equation} \label{eq:wt_tree}
%\mathbf{w}_{k}^{i} = \textrm{TreeCell}_{\boldsymbol{\theta}}^{\text{weight}}(\mathbf{w}^{i-1}_{2j-1}, \mathbf{w}^{i-1}_{2j}),
%\end{equation}

%where $1 \le i \le \floor{\log(k - 1)} + 1$, $1 \le j \le \floor{\frac{k}{2^{i}}}$,

\paragraph{Edge Weight Conditioning}

Because the summary weight state $\mathbf{h}_k^{wt}$ encodes information about all previously generated weights $w_1$ to $w_k$, using this state to predict the next edge weight $w_{k+1}$ allows BiGG-E to condition each new weight on the history of prior weights. As established in Section~\ref{sec:joint}, each edge weight is sampled from a softplus normal distribution with mean $\mu_k$ and variance $\sigma^2_{k}$ parameterized by functions of $\boldsymbol{\theta}$. Computing these parameters from $\mathbf{h}_{k}^{wt}$ in the sampling of the next weight $w_{k+1}$ allows for conditioning $\mu_{k+1}$ and $\sigma^{2}_{k+1}$ on all preceding weights:

\begin{equation} \label{eq:biggemu}
\begin{aligned}
\mu_{k+1} &= f_{\mu}(\mathbf{h}^{wt}_k)
\end{aligned}
\hspace{1cm}
\begin{aligned}
\log \sigma^{2}_{k+1} &= f_{\sigma^2}(\mathbf{h}^{wt}_k)
\end{aligned}
\end{equation}

where $f_{\mu}$ and $f_{\sigma^2}$ are MLPs that output the estimated mean and log-scale variance, respectively. While this allows BiGG-E to model dependencies among weights, a full generative model must also capture how edge structure and weights influence one another -- a task requiring both the topology and weight states.

\begin{figure} [t]
\centering
\resizebox{0.75\linewidth}{!}{
\begin{tikzpicture}[node distance=0.7cm and 2cm, every node/.style={font=\small}, edge/.style={thick}, dashededge/.style={thick, dashed, red}, level distance=0.5cm, sibling distance=0.75cm]

\begin{scope}[yshift=-3.0cm, xshift=-3.75cm]
\node (title)[] at (-5, 8.125) {Current Graph};
\node (v2)[circle, draw] at (-6,7) {$v_1$};
\node (v1)[circle, draw] at (-6,5) {$v_2$};
\node (v4)[circle, draw] at (-4,5) {$v_3$};
\node (v3)[circle, draw] at (-4,7) {$v_4$};
\node (v5) at (-3.5,7.5) {};

% Solid edges
\draw[edge] (v1) -- node[left]  {$w_1$} (v2);
\draw[edge] (v2) -- node[above] {$w_4$} (v3);
\draw[edge] (v3) -- node[right] {$w_3$} (v4);
\draw[edge] (v1) -- node[below] {$w_2$} (v4);

\end{scope}

% ---------- Topology Tree (Narrow + Vertically Aligned Leaves) ----------
\begin{scope}[xshift = 0cm]
\node[circle, draw, solid, fill=yellow, minimum size=0.3cm, inner sep=1pt] (paramsk) at (0.3,-0.5) {};
\node[circle, draw, solid, fill=yellow, minimum size=0.3cm, inner sep=1pt] (paramsk1) at (4.7,-0.5) {};

% Root node
\node (t0) at ($(paramsk)!0.5!(paramsk1)+(0,6)$) {$[v_1, v_4]$};

% Inner tree nodes
\node (t1) at ($(paramsk)!0.2!(paramsk1)+(0.1,3.8)$) {$[v_1, v_2]$};
\node (t2) at ($(paramsk)!0.8!(paramsk1)+(-0.1,3.8)$) {$[v_3, v_4]$};

% Tree-LSTM Nodes
\node[circle, draw, solid, fill=yellow, minimum size=0.3cm, inner sep=1pt] (l1) at ($(paramsk)!0.3!(paramsk1)+(0.1,4.8)$) {};
\node[circle, draw, solid, fill=yellow, minimum size=0.3cm, inner sep=1pt] (l2) at ($(paramsk)!0.4!(paramsk1)+(1.1,4.8)$) {};
\node[circle, draw, solid, fill=yellow, minimum size=0.3cm, inner sep=1pt] (l3) at ($(paramsk)!0.5!(paramsk1)+(-1.8,2.75)$) {};
\node[circle, draw, solid, fill=yellow, minimum size=0.3cm, inner sep=1pt] (l4) at ($(paramsk)!0.6!(paramsk1)+(1.35,2.75)$) {};

% Leaf nodes
\node (vk) at ($(paramsk)+(0,1.4)$) {$v_1$};
\node (vk1) at ($(paramsk1)+(0,1.4)$) {$v_4$};

% Phantom points for rightward stretch (green box)
\node (phantomRightTop) at ($(t0)+(0.50,-4.35)$) {};
\node (phantomRightBot) at ($(vk1)+(0.50,0.5)$) {};
\node (phantomLeftBot) at ($(t0)+(-2.70,0)$) {};
\node (phantomLeftTop) at ($(vk1)+(-2.70,5.05)$) {};

% Tree edges
\draw[->, dashed, thick, blue] (t0) -- (l1);
\draw[->, thick] (l1) -- (t1);
\draw[->, dashed, thick, blue] (t0) -- (l2);
\draw[->, thick] (l2) -- (t2);
\draw[->, dashed, thick, blue] (t1) -- (l3);
\draw[->, thick] (l3) -- (vk);
\draw[->, dashed, thick, blue] (t2) -- (l4);
\draw[->, thick] (l4) -- (vk1);

% Framing box (extended right)
%\begin{pgfonlayer}{background}
%  \node[draw=blue!60, thick, rounded corners, fit=(t0)(t1)(t2)(vk)(vk1)(phantomRightTop)(phantomRightBot)(phantomLeftBot)(phantomLeftTop), inner sep=0.25cm] {}; %label=above:{\textbf{Construction of $\mathcal{T}_{v_5}$}}] {};
%\end{pgfonlayer}

% ---------- Weight Sampling + Embedding Block ----------
\begin{scope}[yshift=-2.5cm] %This moves LSTM to H block

% Weight sampling
\node[below=1cm of paramsk] (wk) {$w_5$};
\node[draw, rectangle, minimum width=1.2cm, minimum height=0.8cm, below=0.6cm of wk] (lstm_k) {LSTM};

\node[below=1cm of paramsk1] (wk1) {$w_6$};
\node[draw, rectangle, minimum width=1.2cm, minimum height=0.8cm, below=0.6cm of wk1] (lstm_k1) {LSTM};

% Hidden states
\node[draw, rectangle, minimum width=1.2cm, minimum height=1cm] (hk) at (0.3,-2.5) {$\mathbf{h}_5^{wt}$};
\node[draw, rectangle, minimum width=1.2cm, minimum height=1cm, right=of hk, xshift=1.2cm] (hk1) {$\mathbf{h}_{6}^{wt}$};
\node (hkm1) {};

% Phantom points for leftward stretch (blue box)
\node (phantomLeftTop) at ($(hkm1)+(.18,  0.2)$) {};
\node (pantomRightTop) at ($(hkm1)+(5.0, 0.35)$) {};

% Arrows
\draw[->, thick] (paramsk) -- (wk);
\draw[thick, purple] (wk) -- (lstm_k);
\draw[->, thick, purple] (lstm_k) -- ($(hk.north)+(0,-0.05)$);

\draw[->, thick] (paramsk1) -- (wk1);
\draw[thick, purple] (wk1) -- (lstm_k1);
\draw[->, thick, purple] (lstm_k1) -- ($(hk1.north)+(0,-0.05)$);

\draw[->, thick, purple] (hk.east) -- (hk1.west);

\draw[->, thick, blue, dashed] (vk.south) -- (paramsk.north);
\draw[->, thick, blue, dashed] (vk1.south) -- (paramsk1.north);
\draw[->, thick, green!60!black, dashed] (hk.north east) to[bend left=20] (paramsk1.west);

\draw[->, thick, green!60!black, dashed] (hk.north east) to[bend left=5] (l2.west);
\draw[->, thick, green!60!black, dashed] (hk.north east) to[bend left=10] (l4.west);

% Framing box (extended left)
%\begin{pgfonlayer}{background}
%  \node[draw=green!50!black, thick, rounded corners, fit=(hk)(hk1)(pantomRightTop)(phantomLeftTop), inner sep=0.33cm, label=below:{\textbf{Weight Embedding Block}}] {};
%\end{pgfonlayer}

\end{scope}

\node (phantomBL) at ($(hkm1)+(-0.4,  -3.7)$) {};
\node (phantomBR) at ($(hkm1)+(5.55, 1.2)$) {};
\node (phantomUL) at ($(hkm1)+(-0.4,  8.68)$) {};

\begin{pgfonlayer}{background}
  \node[draw=black, thick, rounded corners, fit=(t0)(t1)(t2)(vk)(vk1)(phantomRightTop)(phantomRightBot)(phantomLeftBot)(phantomLeftTop)(paramsk)(paramsk1)(hk)(hk1)(pantomRightTop)(phantomLeftTop)(phantomBL)(phantomBR)(phantomUL), inner sep=0.25cm, label=above:{\textbf{(B) Construction of $\mathcal{T}_{v_5}$}}] {};
\end{pgfonlayer}

% ------------------ Graph Summary Box ------------ %
\begin{scope}[xshift = -2.5cm, yshift = 2cm]

\node[draw, rectangle, minimum width=1.2cm, minimum height=1cm, xshift=0.3cm] (hk4) {$\mathbf{h}_{4}^{wt}$};
\node[draw, rectangle, minimum width=1.2cm, minimum height=1cm, xshift=0.3cm, yshift=2.75cm] (hrow4) {$\mathbf{h}_{4}^{row}$};
\node (title)[xshift=-2.5cm, yshift= 4.25cm] {$\textbf{Fenwick Trees}$};
\node[xshift=5.14cm, yshift = 4.24cm] (test) {};
\node (rowt)[xshift=-2.5cm, yshift =3.25cm] {Topology};
\node (wtt)[xshift=-2.5cm, yshift =1cm] {Weight};

\draw[->, thick, green!60!black, dashed] (hk4.east) to[bend left=-10] (paramsk.west);
\draw[->, thick, green!60!black, dashed] (hk4.east) to[bend left=22] (l1.west);
\draw[->, thick, green!60!black, dashed] (hk4.east) to[bend left=-2] (l3.west);
\draw[thick, purple] (hk4.south) |- (hk.west);
\draw[->, thick, blue, dashed] (hrow4.east) to[bend left=20] (test) to (t0.north);

% Define the Fenwick Tree
\node[circle, fill=black, inner sep=1pt, xshift=-2.5cm, yshift=2.75cm] (tr1) {}
  child {node[circle, fill=black, inner sep=1pt] {}
    child {node[circle, fill=black, inner sep=1pt, xshift = 0.15cm] {}}
    child {node[circle, fill=black, inner sep=1pt, xshift = -0.15cm] {}}
  }
  child {node[circle, fill=black, inner sep=1pt] {}
    child {node[circle, fill=black, inner sep=1pt, xshift = 0.15cm] {}}
    child {node[circle, fill=black, inner sep=1pt, xshift = -0.15cm] {}}
  };

% Define the Weight Fenwick Tree
\node[circle, fill=black, inner sep=1pt, xshift=-2.5cm, yshift=0.5cm] (tr2) {}
  child {node[circle, fill=black, inner sep=1pt] {}
    child {node[circle, fill=black, inner sep=1pt, xshift = 0.15cm] {}}
    child {node[circle, fill=black, inner sep=1pt, xshift = -0.15cm] {}}
  }
  child {node[circle, fill=black, inner sep=1pt] {}
    child {node[circle, fill=black, inner sep=1pt, xshift = 0.15cm] {}}
    child {node[circle, fill=black, inner sep=1pt, xshift = -0.15cm] (lch) {}}
  };

\node (phantom1) at (-3.4, 0.0) {};
\node (phantom2) at (-1.5, 0.0) {};

\begin{pgfonlayer}{background}
  \node[draw=black, thick, rounded corners, dashed, fit=(phantom1)(phantom2)(rowt)(wtt)(tr1)(lch), inner sep=0.33cm] {};
\end{pgfonlayer}

\draw[->, thick, color=blue, draw opacity=0.8, dashed] (tr1) to[bend right = 20]  ($(hrow4.west) + (0, -0.22)$);
\draw[->, thick, color=green!60!black, draw opacity=0.8, dashed] (tr2) to[bend right = 10] (hk4.west);

\node (phantomBR) at ($(hk4)+(0.68,-0.77)$) {};
\node (phantomUL) at ($(hk4)+(-4.25,4.1)$) {};

\begin{pgfonlayer}{background}
  \node[draw=black, thick, rounded corners, fit=(v2)(hk4)(phantomUL)(phantomBR), inner sep=0.33cm, label=above:{\textbf{(A) Graph Summary + Initial States}}] {};
\end{pgfonlayer}
\end{scope}

% ---------- Complete Graph ----------
\begin{scope}[yshift=-9cm, xshift=-3.75cm]
\node (v2)[circle, draw] at (-6,5) {$v_1$};
\node (v1)[circle, draw] at (-6,3) {$v_2$};
\node (v4)[circle, draw] at (-4,3) {$v_3$};
\node (v3)[circle, draw] at (-4,5) {$v_4$};
\node (v5)[circle, draw, dashed, red] at (-5,7) {$v_5$};

\node (phantomTL) at (-6.25,8.4) {};
\node (phantomBR) at (0.8,2.87) {};

\node (tv5) at (-1, 7.9) {$\mathcal{T}_{v_{5}}$};
\node (ng) at (-5, 7.9) {Updated Graph};

% Root node
\node (tt0) at (-1, 7) {$[v_1, v_4]$};
\node (tt1) at (-1.75, 5) {$[v_1, v_2]$};
\node (tt2) at (-0.25, 5) {$[v_3, v_4]$};
\node (tt3) at (-2.25, 3) {$(v_1, w_1)$};
\node (tt4) at (0.25, 3) {$(v_4, w_4)$};

% Tree Lines
\draw[thick] (tt0) -- (tt1);
\draw[thick] (tt0) -- (tt2);
\draw[thick] (tt1) -- (tt3);
\draw[thick] (tt2) -- (tt4);

% Solid edges
\draw[edge] (v1) -- node[left]  {$w_1$} (v2);
\draw[edge] (v2) -- node[above] {$w_4$} (v3);
\draw[edge] (v3) -- node[right] {$w_3$} (v4);
\draw[edge] (v1) -- node[below] {$w_2$} (v4);

% Dashed edges to v5
\draw[dashededge] (v2) -- node[left=2pt] {$w_5$} (v5);
\draw[dashededge] (v3) -- node[right=2pt] {$w_6$} (v5);

\begin{pgfonlayer}{background}
  \node[draw=black, thick, rounded corners, fit=(phantomTL)(phantomBR), inner sep=0.33cm, label=above:{\textbf{(C) Adding edge connections to $v_5$}}] {};
\end{pgfonlayer}

\end{scope}

% ---------- Legend / Key ----------
\begin{scope}[shift={(-9.24,-9.7)}]

\draw[->, thick, blue, dashed] (-1,2) -- (-0.4,2);
\node[anchor=west] (legendBlue) at (-0.2,2) {Incoming};
\node[anchor=west] (legendBlue2) at (-0.2,1.5) {Topological State};

\draw[->, thick, green!60!black, dashed] (3,2) -- (3.6,2);
\node[anchor=west] (legendRed) at (3.8,2) {Incoming};
\node[anchor=west] (legendRed2) at (3.8,1.5) {Weight State};

\node[circle, draw, solid, fill=yellow, minimum size=0.3cm, inner sep=1pt] at (6.5, 2) {};
\node[anchor=west] (legendTree) at (6.7, 2) {Tree-LSTM};
\node[anchor=west] (legendTree2) at (6.7, 1.5) {Cell};

\draw[->, thick, black] (9.1,2) -- (9.7,2);
\node[anchor=west] (legendBlack) at (9.8,2) {Merged};
\node[anchor=west] (legendBlack2) at (9.8,1.5) {Prediction};

\node (legendTitle) at (-0.7,2.2) {};
\draw[->, thick, purple] (11.7,2) -- (12.3,2);
\node[anchor=west] (legendPurple) at (12.3,2) {Weight};
\node[anchor=west] (legendPurple2) at  (12.3,1.5) {Tree Update};

\node (Phantom) at (14.64, 1.5) {};

% Dashed box around legend
\begin{pgfonlayer}{background}
  \node[draw=gray, thick, rounded corners, dashed, fit=(legendTitle)(legendPurple)(legendPurple2)(legendRed)(legendRed2)(legendBlue)(legendBlue2)(Phantom), inner sep=0.4cm] {};
\end{pgfonlayer}
\end{scope}
\end{scope}

\end{tikzpicture}
}
\captionof{figure}[Example of Autoregressive Construction of $\mathcal{T}_{v_5}$ with Edge Weights.]
{Illustration of the autoregressive construction of $\mathcal{T}_{v_5}$ with weighted edges. Blue dashed arrows indicate the topological state; green dashed arrows indicate the weight state. These are merged with Tree-LSTM cells (yellow) to output hidden states for predicting new weighted edges. New weights are embedded using an LSTM and then added to the Fenwick weight tree. Panels (A–C) depict the process from graph summarization to final integration of $v_5$ into the graph.}
\label{fig:embedfig}
\centering
\end{figure}
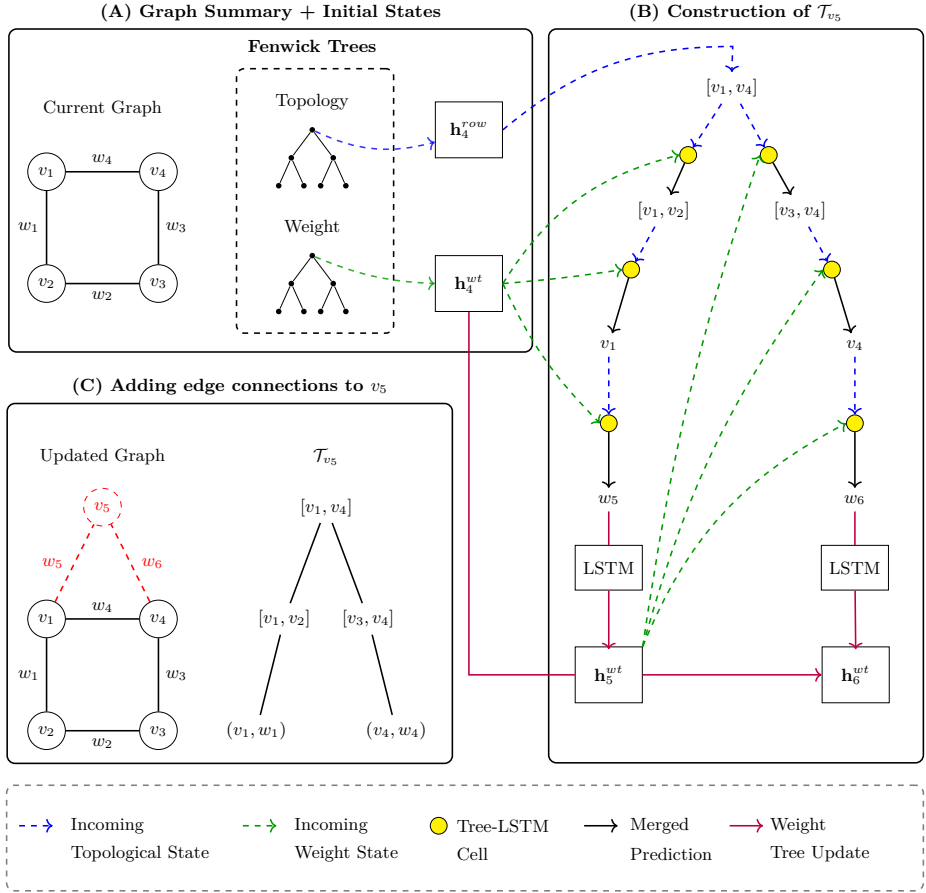

\subsubsection{Joint Modeling} \label{sec:jointbigge}
Currently, the topology state is responsible for conditioning edge formation, while the weight state conditions edge weights. Used independently, each state informs only its respective component.  However, by leveraging both states during prediction, BiGG-E models the joint interaction between topology and weights. Let $\mathbf{h}^{\text{top}}_{u}(t)$ be the current top-down context at node $t \in \mathcal{T}_{u}$ and $\mathbf{h}^{\text{wt}}_{k}$ be the current weight state of the most recent weight $w_k$. BiGG-E computes the weighted graph summary state $\mathbf{h}^{\text{sum}}_{u,k}(t)$ using a Tree-LSTM cell:

\begin{equation} \label{eq:merge}
\mathbf{h}^{\text{sum}}_{u,k}(t) = \text{TreeCell}^{\text{merge}}_{\boldsymbol{\theta}}(\mathbf{h}^{\text{top}}_{u}(t), \mathbf{h}^{\text{wt}}_{k}).
\end{equation} 

First, when constructing $\mathcal{T}_u$, BiGG-E merges $\mathbf{h}^{\text{wt}}_{k}$ with the top-down context vectors $\mathbf{h}_{u}^{\text{top}}(t)$ and $\mathbf{\hat{h}}_{u}^{\text{top}}(t)$ before making predictions for lch($t$) and rch($t$), respectively, using Equation~\ref{eq:merge}. This modifies Equation~\ref{eq:tuprobbigg} to 

\[
\begin{aligned} \label{eq:jointmodel}
p_{\boldsymbol{\theta}}(\mathcal{T}_u ) =  \prod_{t\in \mathcal{T}_u} & p_{\boldsymbol{\theta}}(\textrm{lch}(t) | \mathbf{h}_{u,k}^{\text{sum}}(t)) \cdot p_{\boldsymbol{\theta}}(\textrm{rch(t)} | \mathbf{\hat{h}}_{u,k}^{\text{sum}}(t)),
\end{aligned}
\]

which enables further conditioning the graph's topology on the corresponding edge weights.

Next, when BiGG-E forms an edge connection at a singleton interval $t = [v_j, v_j]$ in $\mathcal{T}_u$, the model merges $\mathbf{h}_{u}^{\text{top}}(t)$ with $\mathbf{h}^{\text{wt}}_{k}$ using Equation~\ref{eq:merge} prior to outputting the mean and variance parameters $\mu_{k+1}$ and $\sigma^2_{k+1}$. This modifies Equation \ref{eq:biggemu} to

\[
\begin{aligned}
\mu_{k+1} &= f_{\mu}(\mathbf{h}_{u, k}^{\text{sum}}(t))
\end{aligned}
\hspace{1cm}
\begin{aligned}
\log \sigma^{2}_{k+1} &= f_{\sigma^2}(\mathbf{h}_{u, k}^{\text{sum}}(t))
\end{aligned}
\]
which conditions sampling edge weights on the topology of the graph, fully allowing BiGG-E to jointly model weighted graphs. Algorithm~\ref{alg1} outlines sampling a weight when an edge exists, and Figure~\ref{fig:embedfig} visualizes the joint modeling of weighted edges when constructing $\mathcal{T}_u$.

\subsubsection{BiGG-E Training and Sampling Time}

The novel architectural design outlined in Section~\ref{sec:bigge-desc} allows BiGG-E to extend the efficient training and sampling observed from BiGG to sparse weighted graphs. Constructing the weight state with the Fenwick weight tree adds only $\mathcal{O}(\log m)$ training time and $\mathcal{O}(m \log m)$ sampling time, which preserves the overall asymptotic complexity of the model so long as $m = \mathcal{O}(n)$.  Furthermore, since BiGG-E constructs each tree $\mathcal{T}_u$ using BiGG's original procedure, training remains parallelizable across rows. Finally, the use of Tree-LSTM cells to merge topological and weight states does not disrupt this structure, allowing BiGG-E to maintain a training time of $\mathcal{O}(\log n)$ and sampling time of $\mathcal{O}((n + m)\log n)$ for sparse weighted graphs.

\subsection{Comparison Models}

Our code for BiGG-E and all comparison models is available at \href{https://github.com/rlwilliams34/BiGG-E}{https://github.com/rlwilliams34/BiGG-E}.

\paragraph{Adj-LSTM}

Adj-LSTM builds on the work of \citet{li} by using an LSTM cell to parameterize the lower half of $\mathbf{W}$ in a row-wise fashion. To adapt to the grid-like structure of $\mathbf{W}$, Adj-LSTM maintains and updates row and column states autoregressively. When an edge exists between nodes $v_i$ and $v_j$, a corresponding weight is sampled as described in Section~\ref{sec:joint}. Adj-LSTM is fully expressive but slow: generating all of $\mathbf{W}$ requires $\mathcal{O}(n^2)$ computations. Additional details are provided in Appendix~\ref{sec:adjlstm}.

\paragraph{BiGG-MLP}

BiGG-MLP is a simple extension of BiGG that replaces the topology leaf state (Algorithm ~\ref{alg1}, line 9) that indicates edge existences in $\mathcal{T}_{u}$ with an MLP that encodes the newly sampled weight into a state embedding. The rest of the algorithm remains unchanged, maintaining a single state that summarizes topology and edge weights. However, using the same state to simultaneously make left-right binary decisions and generate continuous weights severely limits BiGG-MLP's capacity to learn each task. This comparison highlights the importance of maintaining separate topology and weight states, as done in BiGG-E.

\paragraph{BiGG+GCN}

BiGG+GCN is a two-stage model that decouples topology generation from edge weight sampling. First, the original BiGG model generates an unweighted graph. Then, a graph convolutional network (GCN) conditioned on the graph's topology populates each edge with edge weights. Thus, BiGG+GCN forgoes jointly modeling weighted edges and instead generates edges independently of weights. BiGG+GCN serves as a comparison for evaluating the benefits of BiGG-E's joint modeling approach.

\paragraph{Baseline}

The baseline is an Erdős–Rényi (ER) model that estimates a global edge existence probability and samples each edge independently. Corresponding weights are sampled with replacement from the distribution of edge weights in the training graph set.

\begin{table*}[t]
  \centering
    \caption{Topological Accuracy Measures. The MMD metrics use the test functions from the set \{Degree, Cluster, Orbit, Unweighted Laplacian (Spec)\}. For the MMD metrics, smaller values are better. OOM indicates out of memory. Error is reported as the proportion of non-tree or -lobster graphs. Similar or better topological accuracy compared to the original BiGG (BiGG+GCN) shows that modeling of weights does not worsen BiGG-E's topological performance.}
  \begin{tabular}{ccccccc}
    \specialrule{1.25pt}{1pt}{1pt} \\[-2mm] 
    \specialrule{0.75pt}{1pt}{1pt}
    \multirow{2.5}{*}{Datasets} & \multicolumn{6}{c}{Methods} \\
    \cmidrule{3-7} & {} & BiGG-E & Adj-LSTM &  BiGG-MLP & BiGG+GCN & Erdős–Rényi\\
    \midrule
    \multirow{2}{*}{Erdős–Rényi} & Deg. & $\boldsymbol{1.75e^{-3}}$  & 0.437 & $1.11e^{-2}$ & $3.76e^{-3}$ & $2.80e^{-3}$ \\
    {} & Clus.  & $9.35e^{-3}$ & 0.744 &  $1.27e^{-2}$ & $\boldsymbol{7.50e^{-3}}$ & $1.48e^{-2}$
    \\
    $|V|_{max} = 749$ $(499)$ & Orbit &  $6.11e^{-2}$ & 0.263 & $6.66e^{-2}$ & $\boldsymbol{6.08e^{-2}}$ & $8.31e^{-2}$
    \\
    $|E|_{max} = 2846$ $(1349)$ & Spec. & $2.92e^{-3}$  & 0.516  & $9.61e^{-3}$ & $2.71e^{-3}$ & $\boldsymbol{2.06e^{-3}}$
    \\
    \midrule
    \multirow{2}{*}{Tree} & Deg. & $\boldsymbol{2.94e^{-6}}$ & $8.31e^{-4}$ & $1.54e^{-4}$ & $8.47e^{-6}$ & $0.270$
    \\
    {}  & Spec. & $\boldsymbol{3.94e^{-4}}$ & $2.27e^{-3}$  &  $2.63e^{-3}$ & $6.11e^{-4}$ & $8.31e^{-2}$
    \\
    $|V|_{max} = 199$ $(199)$ & Orbit & $6.46e^{-9}$  & $8.15e^{-6}$ &  $3.49e^{-6}$ & $\boldsymbol{4.69e^{-9}}$ & $3.90e^{-2}$
    \\
    $|E|_{max} = 198$ $(198)$ & Error & $\boldsymbol{0.025}$  & $0.285$ & $0.76$ & $0.065$ & $1.0$
    \\
    \midrule
    \multirow{2}{*}{3D Point Cloud} & Deg. & $7.40e^{-3}$ & OOM  & 0.247 & $\boldsymbol{4.37e^{-4}}$ & $0.477$
    \\
    {} & Clus.  & $0.179$  & OOM & $0.479$ &  $\boldsymbol{0.175}$ &   $1.143$
    \\
    $|V|_{max} = 5022$ $(1375)$ & Orbit& $5.06e^{-3}$ & OOM  & $0.224$ & $\boldsymbol{1.15e^{-3}}$ & $0.960$
    \\ 
    $|E|_{max} = 10794$ $(3061)$ & Spec. & $\boldsymbol{7.40e^{-3}}$ & OOM  & $2.82e^{-2}$ & $7.50e^{-3}$ & $0.144$
    \\
    \midrule
    \multirow{2}{*}{Lobster} & Deg.  & $\boldsymbol{4.32e^{-4}}$  & $4.63e^{-4}$& $2.33e^{-3}$ & $1.18e^{-3}$ & $0.173$ \\
    {} & Clus. & $\boldsymbol{0.0}$ & $1.89e^{-4}$  & $\boldsymbol{0.0}$  & $\boldsymbol{0.0}$ & $\boldsymbol{0.0}$
    \\
    $|V|_{max} = 100$ $(55)$ & Spec.  & $3.14e^{-3}$  & $\boldsymbol{2.50e^{-3}}$ & $4.48e^{-3}$ & $3.54e^{-3}$ & $0.189$
    \\
    $|E|_{max} = 99$ $(54)$ & Orbit  & $\boldsymbol{3.25e^{-3}}$ &  $6.36e^{-3}$ & $3.38e^{-2}$ & $9.64e^{-3}$ & $0.167$
    \\
    {} & Error & $\boldsymbol{0.005}$ & $0.195$  & $0.315$ & $0.165$ & $1.0$
    \\
    \bottomrule
  \end{tabular}
  \label{tab:tab1}
\end{table*}

\begin{table*}[t]
  \centering
    \captionof{table}[Edge-weight Accuracy Measures]{Edge-weight Accuracy Measures. The MMD metrics use the test functions from the set \{Weighted Laplacian (Spec) and Weights\}. For the MMD metrics, smaller values are better. OOM indicates out of memory. First and second order statistics (Mean and SD) of test sets are given. For ER Graphs, $\mu_z$ and $\sigma_z$ are the mean and SD of the standard normal distribution. Sample means and SDs are back-transformed. For tree graphs, $w$ represents global weight estimates and $T$ represents per-tree estimates. Similar or better weight accuracy compared to Adj-LSTM shows that extending BiGG does not worsen BiGG-E's weight performance.}
  %\textbf{Weight Measures} \\
  \begin{tabular}{ccccccc}
    \specialrule{1.25pt}{1pt}{1pt} \\[-2mm] 
    \specialrule{0.75pt}{1pt}{1pt}
    \multirow{2.5}{*}{Datasets} & \multicolumn{6}{c}{Methods} \\
    \cmidrule{3-7} & {} & BiGG-E & Adj-LSTM & BiGG-MLP & BiGG+GCN & Erdos Renyi\\
    \midrule
    \multirow{2}{*}{Erdős–Rényi} & Mean & $8.17e^{-3}$ & $-0.163$  & $7.53e^{-3}$ & $-1.01e^{-2}$ & $\boldsymbol{2.07e^{-3}}$  \\
    {} & SD & $1.009$  & $1.983$  & $1.012$ & $\boldsymbol{1.002}$  & $0.997$ 
    \\ 
    $\mu_{z} = 0.0$ & Spec. & $4.24e^{-3}$  & $0.469$ & $8.76e^{-3}$ & $3.87e^{-3}$ & $\boldsymbol{1.68e^{-3}}$ 
    \\
    $\sigma_{z} = 1.0$ & MMDWT & $6.83e^{-3}$  & $9.40e^{-3}$& $1.12e^{-2}$ & $3.01e^{-3}$ & $\boldsymbol{2.66e^{-3}}$
    \\
    \midrule
    \multirow{2}{*}{Tree} & $\bar{w}$ & $10.136$ & $\boldsymbol{9.988}$  & $11.136$ & $10.034$  & $10.037$ 
    \\
    {} & $s_{w}$ &  $1.855$  & $\boldsymbol{1.972}$ & $2.073$ & $1.956$ & $1.956$ 
    \\
    $\mu_{w} = 10, \sigma_{w} = 2$ & $s_{T}$ & $\boldsymbol{0.999}$ & $1.107$   & $1.071$ & $1.956$ & $1.963$  
    \\ 
    $\sigma_{T} = 1$ & Spec. & $\boldsymbol{5.43e^{-4}}$ & $1.58e^{-3}$  & $1.54e^{-3}$ & $2.09e^{-3}$ & $0.112$
    \\
    {} & MMDWT  & $7.61e^{-3}$ & $\boldsymbol{1.88e^{-3}}$ &  $0.108$ & $0.216$  & $0.215$  
    \\
    \midrule
    \multirow{2}{*}{3D Point Cloud} & Mean & $\boldsymbol{0.416}$ & OOM  & $0.426$ & $0.426$ & $0.419$  \\
    {} & SD  & $\boldsymbol{0.097}$ & OOM   & $0.103$ & $0.191$  &  $0.099$
    \\ 
    $\bar{w} \approx 0.411$ & Spec. & $7.44e^{-3}$ & OOM  & $2.30e^{-2}$  & $\boldsymbol{7.20e^{-3}}$ &  $0.169$
    \\
    $s_{w} \approx 0.096$ & MMDWT & $\boldsymbol{3.00e^{-3}}$ & OOM   & $4.14e^{-2}$ & $0.118$ & $2.22e^{-2}$
    \\
    {} & Wtd. Deg. & $\boldsymbol{1.84e^{-3}}$ & OOM  & $0.130$  & $2.88e^{-2}$ & $1.01$
    \\
    \midrule
    \multirow{2}{*}{Lobster} & $\bar{w}$ & $0.249$ & $0.249$   & $0.248$ & $0.252$ & $\boldsymbol{0.250}$ \\
    {} & $s_{w}$ & $\boldsymbol{0.094}$ & $0.100$  & $\boldsymbol{0.094}$  & $0.093$  & $0.098$ 
    \\ 
    $\mu = 0.25$ & Spec.  & $2.61e^{-3}$  & $\boldsymbol{2.34e^{-3}}$ & $3.16e^{-3}$  & $2.93e^{-3}$ & $0.259$
    \\
    $\sigma \approx 0.095$ & MMDWT & $\boldsymbol{1.23e^{-3}}$ & $6.44e^{-3}$  & $1.12e^{-3}$ & $1.24e^{-3}$ & $2.60e^{-3}$
    \\
    \bottomrule
    \end{tabular}
  \label{tab:tab2}
\end{table*}

\section{Experiment}

We assess all models on the following: (1) do the models capture diverse distributions over weighted graphs; (2) does jointly modeling topology and edge weights improve performance; and (3) do the models scale well to large graphs? Following the evaluation protocol of \citet{liao}, we compare graphs generated by each model against a held-out test set.

To assess generative quality, we use metrics that evaluate the marginal distributions of graph topology and edge weights, and metrics that evaluate their joint distribution. For topology, we compute Maximum Mean Discrepancy (MMD) \citep{gretton} using test functions based on degree distributions, clustering coefficients distributions, and the spectrum of the normalized unweighted Laplacian. For the tree and lobster datasets, we also report the error rate as the proportion of generated graphs that do not follow the correct structure. To assess edge weight quality, we report first- and second-order summary statistics, as well as the MMD on marginal edge weight distributions from each graph. To evaluate joint structure, we compute the MMD using test functions based on the weighted Laplacian spectrum and the weighted degree distribution.

To evaluate scalability on larger graphs, we train on sets of 80 weighted trees with sizes from \{100, 200, 0.5K, 1K, 2K, 5K, 10K, 15K\}. Each model is trained on a GV100 GPU with 3.2 GB of memory an single precision performance, and we report the MMD on the normalized Laplacian of the resulting weighted graphs against 20 test graphs. For each trained model, we record the time to sample one graph, the time to complete a forward pass, backward pass, and optimizer step on one training graph, and the memory consumption used per graph during training. We also compare scalability with that of the diffusion model SparseDiff \citep{sparsediff}, hypothesizing that BiGG-E scales more efficiently. Finally, we expect the extended BiGG models to scale better than both Adj-LSTM and SparseDiff, while BiGG-E will retain superior generative quality on large graphs.

\paragraph{Data}
We use the following datasets to evaluate the generative quality of our models.

\begin{itemize}
 \item \textbf{Erdős–Rényi}: 100 graphs that represent a null case to test whether the models capture the distribution of weighted graphs under an Erdős–Rényi model \citep{erdos}. We sample weights independently from the standard normal distribution and transform them with the softplus function.
 \item \textbf{Tree}: 1{,}000 bifurcating trees with hierarchically sampled edge weights: for each tree $T_i$, sample $\mu_i \sim \mathcal{U}(7, 13)$ from the uniform distribution, then $w_{ik} \sim \Gamma(\mu_i^2, \mu_i^{-1})$ from the gamma distribution. This yields a global weight distribution with mean 10 and standard deviation 2, and within-tree standard deviation of 1.
 \item \textbf{3D Point Cloud}: 41 graphs of household objects \citep{neumann}. Weights from the 3D Point Cloud graphs represent the Euclidean distance between the two nodes in each edge.
 \item \textbf{Lobster}: 1{,}000 graphs that are path graphs with edges appended at most two edges away from the backbone. We sample weights independently from the Beta distribution as $w_{k} \sim \textrm{Beta}(5, 15)$. 
 \item \textbf{Joint}: 100 graphs of trees where weights and topology are coupled. We sample weights $w_k \sim \mathcal{U}(0.5, 1.5)$ independently. Starting at the root node, we add edges until the sum of outgoing edge weights exceeds a threshold of 4. For each new child, add edges recursively using the same rule: add children until the total weight of all new edges - plus the path length from the root to the current node - exceeds 4. This process continues until all root-to-leaf path sums exceed 4.
\end{itemize}

\subsection{Results}

\paragraph{Weighted Graph Distributions}

BiGG-E consistently outperforms competing models on topological metrics (Table~\ref{tab:tab1}). With the tree and lobster graphs, it consistently achieves the best or most competitive MMDs (Table~\ref{tab:tab1}). For the Erdős–Rényi graphs, all BiGG extensions are competitive with the baseline, where the models successfully capture a known probability distribution. All models perform equally well, which is expected given the graphs are fully independent. Even on the complex 3D point clouds, BiGG-E closely matches BiGG+GCN, which holds a slight advantage in unweighted degree MMD ($7.40e^{-3}$ vs $4.37e^{-4}$) but a worse Laplacian spetcrum MMD ($7.50e^{-3}$ vs $7.40e^{-3}$). BiGG-E also achieves the lowest error rates for tree and lobster graphs ($2.5\%$ and $0.5\%$, respectively). Finally, BiGG-MLP exhibits substantial degradation in topological quality across graph datasets, performing orders of magnitude worse on tree and 3D point cloud graphs, and showing moderate degradation on lobster graphs.

\begin{figure}[b]
\centering
\includegraphics[width=0.75\linewidth]{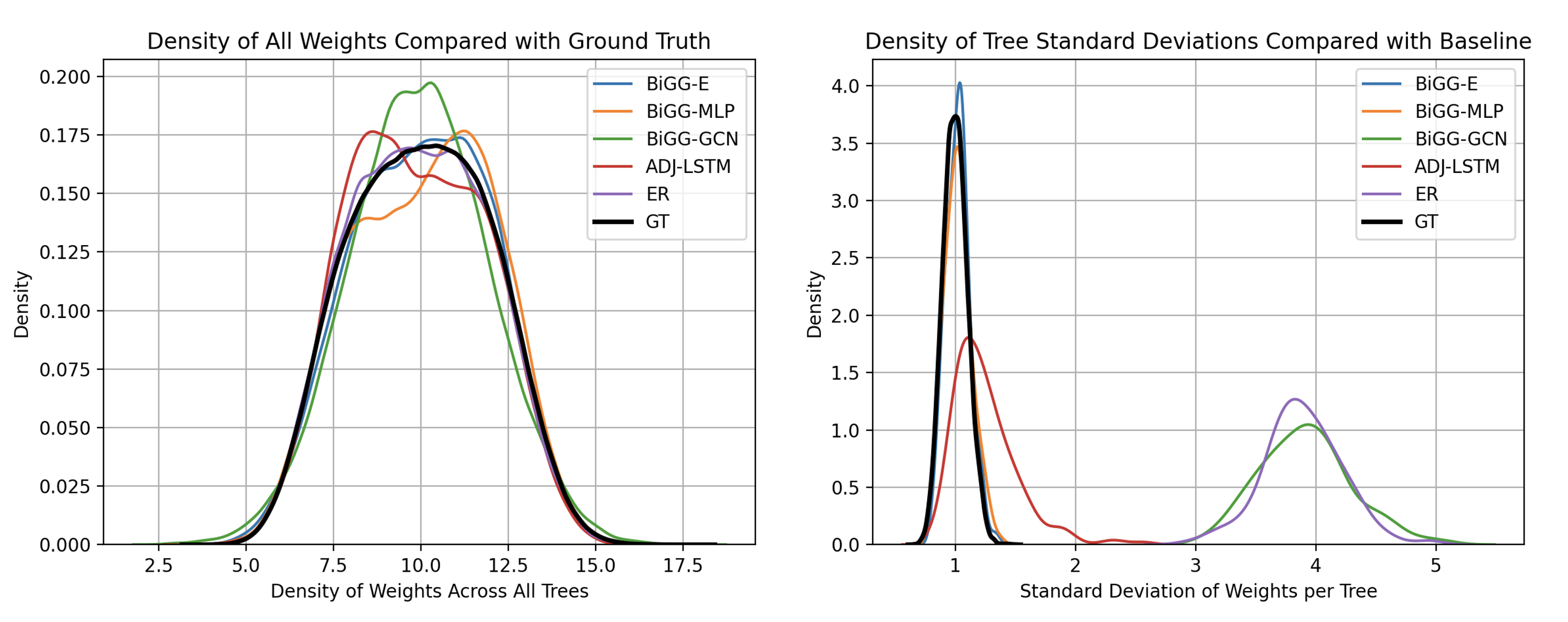}
\captionof{figure}[Tree Weight Distribution Visualization]{Distribution of weights globally (left) and of standard deviations per tree (right).}
\label{fig:tree_wt}
\end{figure}

BiGG-E maintains strong performance in modeling marginal edge weight distributions, particularly with capturing dependencies in the tree and 3D point cloud graphs (Table~\ref{tab:tab2}). Although BiGG+GCN and the Erdős–Rényi baseline perform well on independent weights, they fail in cases where weights are dependent, as seen in the negligible gap between global and per-tree standard deviation on tree graphs. Figure~\ref{fig:tree_wt} reinforces this: BiGG-E and BiGG-MLP align best with the distribution of weights stratified among the tree graphs. On the 3D point cloud graphs, BiGG+GCN generates implausible weights with an inflated standard deviation (0.191 vs. 0.096) despite strong topological performance. Only BiGG-E outperforms the baseline Erdős–Rényi model on weight generation, achieving a significantly lower MMD ($3.00e^{-3}$ vs. $2.22e^{-2}$), and global means and standard deviations that are consistent with the test graphs. 

\begin{table*}[t]
  \centering
    \caption{Topological Accuracy Measures. The MMD metrics use the test functions from the set \{Unweighted and Weighted (Wtd) Degree, Cluster, Orbit, Unweighted and Weighted Laplacian (Spec), Weights (MMDWT)\}. For the MMD metrics, smaller values are better.  Error is the proportion of non-tree graphs.}
  \begin{tabular}{ccccccc}
    \specialrule{1.25pt}{1pt}{1pt} \\[-2mm] 
    \specialrule{0.75pt}{1pt}{1pt}
    \multirow{2.5}{*}{Datasets} & \multicolumn{5}{c}{Methods} \\
    \cmidrule{3-6} & {}  & BiGG-E & BiGG+GCN & BiGG-MLP & Erdős–Rényi\\
    \midrule
    \multirow{2}{*}{Joint Trees} & Deg.  & $\boldsymbol{2.34e^{-5}}$ & $9.31e^{-5}$ &   $8.54e^{-3}$& $6.04e^{-2}$ \\
    {} & Wtd Deg.   & $\boldsymbol{9.19e^{-5}}$ & $7.98e^{-3}$ & $5.83e^{-3}$ & $0.19$
    \\
     & Spec.   &  $\boldsymbol{5.08e^{-3}}$ & $7.45e^{-3} $& $7.27e^{-2}$& $0.22$
    \\
    $|V|_{max} = 452$ $(193)$ & Wtd. Spec.  & $\boldsymbol{9.84e^{-3}}$ & $1.46e^{-2}$  & $2.97e^{-2}$ & $0.21$
    \\
    & Orbit & $\boldsymbol{3.16e^{-5}}$ & $4.12e^{-4}$ & $5.41e^{-3}$ & $1.45e^{-2}$
    \\
    $|E|_{max} = 451$ $(192)$ & MMDWT & $3.68e^{-3}$ & $2.24e^{-2}$ & $7.71e^{-3}$ & $\boldsymbol{3.74e^{-4}}$
    \\
    & Error & $\boldsymbol{0.0}$ & $\boldsymbol{0.0}$ & $\boldsymbol{0.0}$ & $1.0$
    \\
    \bottomrule
  \end{tabular}
  \label{tab:tab44}
\end{table*}

\paragraph{Joint Modeling}

BiGG-E outperforms all comparisons in jointly modeling weighted graphs, particularly on the 3D point cloud graphs (Table~\ref{tab:tab2}) and joint tree graphs (Table ~\ref{tab:tab44}). Although BiGG+GCN achieves a lower unweighted degree distribution MMD for 3D point clouds, BiGG-E attains a lower weighted degree distribution MMD ($1.84e^{-3}$ vs $2.88e^{-2}$). It also matches with BiGG+GCN on the weighted Laplacian MMD ($7.20e^{-3}$ vs $7.44e^{-3}$), where BiGG-E's more realistic edge weights improve joint modeling.

In the joint tree dataset, where topology and edge weights are sampled jointly, BiGG-E achieves the best performance across all metrics by significant margins. Although the Erdős–Rényi baseline performs well on the marginal distribution of weights, which are independently and uniformly distributed, BiGG-E still closely matches this distribution despite using a softplus-parameterized distribution, while also capturing the structural dependencies exhibited in the joint trees that the baseline cannot model.

In contrast to BiGG-E's stable edge weight learning, the two-stage BiGG+GCN pipeline is highly sensitive to errors in the generated topology of the graph, highlighted by the heavily inflated variance of edge weights produced in the 3D point clouds. Uncertainty in the graph topology propagates to the GCN, leading to instability in the sampled set of edge weights. BiGG-E’s joint modeling scheme leads to more stable and realistic weighted graph generation by allowing uncertainty to flow in both directions: uncertainty in topology of the underlying graphs propagates to uncertainty in the edge weight distribution, and vice versa.

\begin{figure*}[t] 
  \centering
    \captionof{table}{Model Scaling on Trees of Varying Size. Weighted Laplacian MMD is reported. Graphs are weighted trees of increasing size. Lower is better. OOM indicates Out of Memory.}
  \begin{tabular}{cccccccc}
    \specialrule{1.25pt}{1pt}{1pt} \\[-2mm] 
    Model & 100 & 0.5K & 1K & 2K & 5K & 10K & 15K\\
    \midrule
    Erdős–Rényi & 0.073 & 0.103 & 0.114 & 0.122 & 0.128 & 0.131 & 0.134
    \\
    BiGG-E & $2.77e^{-3}$  & $\boldsymbol{2.65e^{-3}}$ & $\boldsymbol{1.31e^{-3}}$ & $\boldsymbol{5.60e^{-4}}$ & $\boldsymbol{3.13e^{-4}}$ & $\boldsymbol{6.00e^{-4}}$ & $\boldsymbol{5.12e^{-4}}$
    \\
    BiGG-MLP & $\boldsymbol{1.36e^{-3}}$ & $4.82e^{-3}$  & $3.18e^{-3}$ & $2.04e^{-3}$ & $1.44e^{-3}$ & $6.32e^{-4}$ & $5.47e^{-4}$
    \\
    Adj-LSTM & $6.11e^{-3}$  & $9.27e^{-3}$  & OOM & OOM & OOM & OOM & OOM
    \\
    BiGG+GCN & $3.67e^{-3}$ & $6.01e^{-3}$ & $2.78e^{-3}$ & $1.81e^{-3}$ & $2.96e^{-3}$ & $4.31e^{-3}$ & $1.74e^{-3}$
    \\
    \bottomrule
  \end{tabular}
  \label{tab:tab3}

\begin{minipage}[t]{0.98\linewidth}
\includegraphics[width=\linewidth]{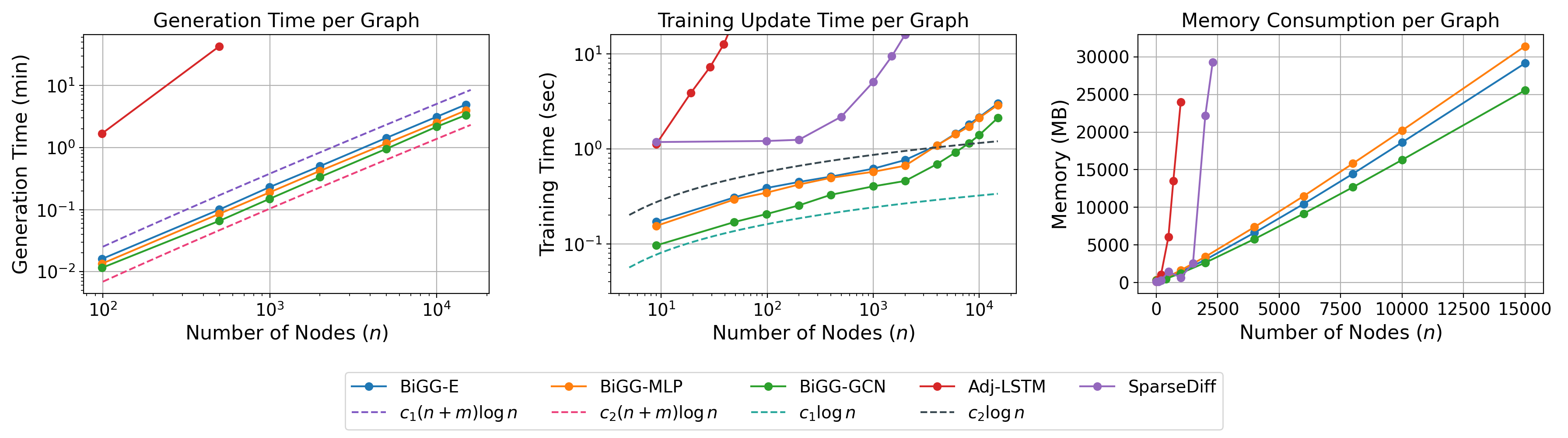}
\captionof{figure}{Model Scalability. Sampling time per weighted graph (left); training time per weighted graph (middle); and memory consumption per graph during training (right).}
\end{minipage}
\label{fig:scale_fig}
\end{figure*}

\paragraph{Scalability}

Table~\ref{tab:tab3} shows that all BiGG extensions scale well to larger graphs, with BiGG-E performing best. Adj-LSTM's performance deteriorates rapidly even on the moderately sized Erdős–Rényi graphs and becomes computationally infeasible for the 3D point clouds, as observed in Table~\ref{tab:tab1}.  Furthermore, Table~\ref{tab:tab3} shows Adj-LSTM fails in scaling to graphs beyond even 500 nodes, where runtime becomes prohibitively slow. On the other hand, all BiGG extensions scale to graphs with thousands of nodes. In Figure~\ref{fig:scale_fig}, we empirically demonstrate that all BiGG extensions remain efficient for large graph generation, with training time scaling as $\mathcal{O}(\log n)$ and sampling time as $\mathcal{O}((n+m)\log n)$, while the training times for Adj-LSTM and SparseDiff increase rapidly with respect to graph size and quickly become impractical. Although BiGG-E's incorporation of a separate weight state slightly increases training and sampling time, the overhead is minimal and justifiable with the superior generative quality. 

Notably, separating the topology and weight states allows BiGG-E to improve memory efficiency. The original BiGG model applies bits compression to summarize the node intervals in $\mathcal{T}_u$ with binary vectors, significantly reducing neural network computation and memory usage \citep{dai}. While this is not feasible in BiGG-MLP from entangling edge weights with the topology state, incorporating bits compression into BiGG-E's topology state is trivial. Moreover, BiGG-E offers flexibility in the dimensionality of each state: we use embedding dimensions of 256 and 32 for the topology and weight states, respectively. This reduces computational overhead and helps prevent overfitting on weights, where we empirically observe a 20\% reduction in BiGG-E's memory consumption. BiGG-MLP, however, uses dense MLPs to output edge embeddings that must match the topology state's embedding dimension of 256, contributing to unstable training and increased memory use. As a result, BiGG-E matches BiGG-MLP’s training time while consuming less memory and producing higher-quality weighted graphs.

\section{Conclusion and Future Work}

We introduce an autoregressive model that learns complex joint distributions over graphs with edge weights. While Adj-LSTM and both BiGG extensions learn from distributions over smaller graphs, BiGG-E scales best to graphs with thousands of nodes while maintaining strong performance learning joint distributions over graph topologies and edge weights. Future work consists of further improving BiGG-E, especially with respect to memory consumption. In addition, we can further explore the benefits of joint modeling edge weights and topologies by learning joint distributions over topologies and vectors of edge and node attributes, learning conditional distributions over these given node- or edge- related data.  Finally and thanks to its scalability, BiGG-E may also be useful within larger regression models featuring network outcomes.

\section{Acknowledgements} 

We would like to extend our thanks to Dr. Hanjun Dai for his help in extending the code base of BiGG to use BiGG-MLP.

Richard Williams is supported by NSF grant DMS 2236854. Andrew Holbrook is supported by the NSF (DMS 2236854, DMS 2152774) and the NIH (K25 AI153816). This work was made  possible by the support of the Cure Alzheimer’s Fund and the Kavli Foundation.

\newpage

\bibliography{main}

\newpage

\appendix
\section{Appendix}

\subsection{Probability Calculation of a Weighted Adjacency Matrix} \label{sec:probcalc}

Here, we derive the probability of observing $\mathbf{W}$ through (1) all entries in a row-wise manner and (2) its weighted edge set. First, we define $p_{\boldsymbol{\theta}}(\mathbf{A})$ over unweighted graphs and then extend these probabilities to probabilities $p_{\boldsymbol{\theta}}(\mathbf{W})$ over weighted graphs. One method in estimating $p_{\boldsymbol{\theta}}(\mathbf{A})$ is to directly generate all of the lower half of $\mathbf{A}$ in a row-wise fashion. Letting $A_{ij}$ represent the $(i, j)\textrm{-th}$ element of $\mathbf{A}$, we have that

\begin{equation} \label{adjedgewt}
p_{\boldsymbol{\theta}}(\mathbf{A}) = \prod_{i = 1}^{n} \prod_{j = 1}^{i-1} p_{\boldsymbol{\theta}}(A_{ij} | \{A_{kl}\}).
\end{equation}
where $\{A_{kl}\}$ is the set of all entries of $\mathbf{A}$ that come before $A_{ij}$ in the row generation process. 

Models such as \citet{li} and \citet{you} use this direct factorization of the probability of all entries of $\mathbf{A}$, leading to the computational bottleneck of $\mathcal{O}(n^2)$ for a graph with $n$ nodes. BiGG \citep{dai}, on the other hand,  leverages the sparsity of many real-world graphs and directly generates the edge-set of $\mathbf{A}$, leading to the probability

\begin{equation}  \label{biggedge22}
p_{\boldsymbol{\theta}}(\mathbf{A}) = \prod_{i = 1}^{m} p_{\boldsymbol{\theta}}(e_{k}|\{e_{l; l < k}\}).
\end{equation}
where $e_k$ is the $i^{\textrm{th}}$ ordered edge in $G$.

Given the choice of parameterization outlined in Section~\ref{sec:joint}, we modify \eqref{adjedgewt} and \eqref{biggedge22} as follows: given an edge exists between nodes $v_{i}$ and $v_{j}$, sample a corresponding weight $w$ from a conditional distribution parameterized by $p_{\boldsymbol{\theta}}(w|e)$. 

For readability, assume entries $W_{ij}$ are conditioned on all prior entries; that is, $p_{\boldsymbol{\theta}}(W_{ij}) \equiv p_{\boldsymbol{\theta}}(W_{ij} | \{W_{kl}\})$, where $\{W_{kl}\}$ is the set of entries of $\mathbf{W}$ that come prior to entry $W_{ij}$ when traversing the lower half of $\mathbf{A}$ in a row-by-row manner.

For each entry $W_{ij}$, we must decide if an edge connects nodes $v_i$ and $v_j$ by sampling $e_{ij} \sim \textrm{Bernoulli}(p_{ij})$, where $p_{ij}$ is a function of $\boldsymbol{\theta}$, and if so, sample a non-negative weight $w_{ij}$ with probability $p_{\boldsymbol{\theta}}(w_{ij}|e_{ij}=1)$. Thus, we note that the probabilities of a particular entry $W_{ij}$ being 0 or weight $w_{ij}$ are given by

\[
p_{\boldsymbol{\theta}}(W_{ij} = 0) = p_{\boldsymbol{\theta}}(e_{ij} = 0) = 1 - p_{ij}
\]
and
\[
p_{\boldsymbol{\theta}}(W_{ij} = w_{ij}) = p_{\boldsymbol{\theta}}(e_{ij} = 1)p_{\boldsymbol{\theta}}(w_{ij} | e_{ij} = 1) = p_{ij} p_{\boldsymbol{\theta}}(w_{ij}|e_{ij}=1).
\]
We may succinctly represent the probability of a $W_{ij}$ as

\[
p_{\boldsymbol{\theta}}(W_{ij} = w_{ij}) = (1-p_{ij})^{1 - e_{ij}} \big[p_{ij} p_{\boldsymbol{\theta}}(w_{ij}|e_{ij})\big]^{e_{ij}}.
\]
Finally, we define the probability of $\mathbf{W}$ as
\[
p_{\boldsymbol{\theta}}(\mathbf{W}) = \prod_{i=1}^{n} \prod_{j=1}^{i-1}(1-p_{ij})^{1 - e_{ij}} \big[p_{ij} p_{\boldsymbol{\theta}}(w_{ij}|e_{ij})\big]^{e_{ij}}.
\]

Next, we modify \eqref{adjedgewt} to obtain the probability of observing $\mathbf{W}$ over all entries as

\[
p_{\boldsymbol{\theta}}(\mathbf{W}) = \prod_{i = 1}^{n} \prod_{j = 1}^{i-1} p_{\boldsymbol{\theta}}(W(v_{i}, v_{j}) | \{W(v_{k}, v_{l})\}).
\]
On the other hand, because no weight is sampled for non-edge connections, the probability of observing the edge set of $\mathbf{W}$ becomes

\[
p_{\boldsymbol{\theta}}(\mathbf{W}) =  \prod_{i = 1}^{m} p_{\boldsymbol{\theta}}(e_{k}, w_{k}|\{(w_i, e_i)\}_{i < k}),
\]
where we factorize $p_{\boldsymbol{\theta}}(e_{k}, w_{k}|\cdot) = p_{\boldsymbol{\theta}}(e_{k} | \cdot) p_{\boldsymbol{\theta}}(w_{k} | e_{k}, \cdot)$ using \eqref{jfactor}.

Finally, we compute our likelihood objective functions. First, we compute $\mathcal{L}(\boldsymbol{\theta}; \mathbf{W})$ from \eqref{adjwprob}, which is used to train Adj-LSTM. The training objective is the log-likelihood over all entries $\mathbf{W}$, summed over the terms

\[
    \ell_{ij}\big(\boldsymbol{\theta}; (e_{ij}, w_{ij})\big) = (1-e_{ij}) \log (1-p_{ij}) + e_{ij} \log p_{ij} + e_{ij} \log p_{\boldsymbol{\theta}}(w_{ij}),
\]

where $e_{ij} \log p_{\boldsymbol{\theta}}(w_{ij}|e_{ij}) = 0$ if $e_{ij} = 0$, and we substitute Equation~\ref{eq:spnorm} into the expression $p_{\boldsymbol{\theta}}(w_{ij}|e_{ij})$ when $e_{ij} = 1$.  This yields the training objective for Adj-LSTM as 

\begin{equation}\label{eq:adjobj}
    \mathcal{L}(\boldsymbol{\theta}; \mathbf{W}) = \log \prod_{i=1}^{n} \prod_{j=1}^{i-1} p_{\boldsymbol{\theta}}(W_{ij}) = \sum_{i=1}^{n} \sum_{j=1}^{i-1}
    \ell_{ij}(\boldsymbol{\theta}; (e_{ij}, w_{ij})).
\end{equation}

Last, from Equation~\ref{biggwprob}, the objective function for all BiGG extensions is the log likelihood over the weighted edge set, 

\begin{equation} \label{eq:biggobj}
    \mathcal{L}(\boldsymbol{\theta}; \mathbf{W}) = \sum_{k=1}^m \log p_{\boldsymbol{\theta}}(e_{k}|\{(e_{l}, w_{l})_{: l < k}\}) +  \log p_{\boldsymbol{\theta}}(w_{k}|e_k, \{(e_{l}, w_{l})_{: l < k}\}),
\end{equation}

where we substitute Equation~\ref{eq:spnorm} into the expression $p_{\boldsymbol{\theta}}(w_{k}|e_k, \{(e_{l}, w_{l})_{: l < k}\})$.

\subsection{Model Architecture} \label{sec:architect}

Autoregressive models are popular sequential models designed to capture dependencies with prior observations. A classical example in statistics is the \( AR(\rho) \) model, where each observation \( X_t \) is expressed as a linear combination of the previous $\rho$ values in the series: \( X_t = \sum_{i=1}^{\rho} \varphi_i X_{t-i} + \epsilon_t \). Deep autoregressive models extend this idea by introducing nonlinear function approximators, such as neural networks, to better model dependencies in sequential data. In this section, we describe in detail the primary neural network architectures used in building each autoregressive model.

\paragraph{LSTM Architecture} The neural network architecture across all models utilizes LSTM cells, which are recurrent neural networks well-suited for modeling nonlinearities in sequential data in an autoregressive manner \citep{hochreiter}. An LSTM maintains a state, represented as a tuple $(\mathbf{h}_t, \mathbf{c}_t)$ where $\mathbf{h}_t, \mathbf{c}_t \in \mathbb{R}^d$ are the hidden and cell states, respectively, at every time step $t$. Here, $\mathbf{h}_t$ captures the recent memory of the sequence, while $\mathbf{c}_t$ encodes longer-term memory. Given an input $\mathbf{x}_t$ at time step $t$ and the previous state $(\mathbf{h}_{t-1}, \mathbf{c}_{t-1})$, the LSTM updates the state by computing $(\mathbf{h}_t, \mathbf{c}_t)$ as follows \citep{sak}:

\begin{itemize}[leftmargin=1in]
    \setlength\itemsep{0.0em}
    \item $\mathbf{i}_t = \sigma(\mathbf{W}_{x}^{(i)}\mathbf{x}_{t} + \mathbf{W}_{h}^{(i)}\mathbf{h}_{t-1} +\mathbf{b}^{(i)})$
    \item $\mathbf{f}_t = \sigma(\mathbf{W}_{x}^{(f)}\mathbf{x}_{t} + \mathbf{W}_{h}^{(f)}\mathbf{h}_{t-1} +\mathbf{b}^{(f)})$
    \item $\mathbf{g}_t = \tanh(\mathbf{W}_{x}^{(g)}\mathbf{x}_{t} + \mathbf{W}_{h}^{(g)}\mathbf{h}_{t-1} +\mathbf{b}^{(g)})$
    \item $\mathbf{o}_t = \sigma(\mathbf{W}_{x}^{(o)}\mathbf{x}_{t} + \mathbf{W}_{h}^{(o)}\mathbf{h}_{t-1} + \mathbf{b}^{(o)})$
    \item $\mathbf{c}_t = \mathbf{f}_t \odot \mathbf{c}_{t-1} + \mathbf{i}_t \odot \mathbf{g}_t$
    \item $\mathbf{h}_t = \mathbf{o}_t \odot \tanh(\mathbf{c}_t)$
\end{itemize}
where $\sigma$ is the sigmoid function, $\odot$ is the Hadamard product, and all weight matrices and bias terms are trainable parameters. $\mathbf{i}_t$, $\mathbf{f}_t$, $\mathbf{g}_t$, and $\mathbf{o}_t$ represent the input, forget, cell, and output gates, respectively. We will also use the notation $(\mathbf{h}_{t+1}, \mathbf{c}_{t+1}) = \textrm{LSTM}(\mathbf{x}_t, (\mathbf{h}_{t}, \mathbf{c}_{t}))$ to represent these computations. For ease of readability, we suppress subscript $t$ moving forward.

\paragraph{MLP} In the graph generative modeling context, an LSTM cell serves to maintain a history of the graph generated by the model at each time step. At each time step, the model must make a prediction—either regarding topology (i.e., the probability of whether an edge exists) or regarding edge weight parameters (e.g., producing the mean and variance of a softplus-transformed normal distribution). To compute these predictions, the hidden state $\mathbf{h}$ is passed through an MLP. An MLP is a feedforward neural network composed of an input layer, one or more hidden layers, and an output layer, where it captures nonlinear patterns in the data through activation functions applied between layers \citep{rumelhart}. In our architecture, we use the exponential linear unit (ELU) as the activation function:

\[ 
\textrm{ELU}(x) = 
  \begin{cases} 
      x & x \ge 0 \\
      \alpha(e^{x}-1) & x <0 \\
   \end{cases}
\]
where $\alpha$ is a tuned parameter. In our case, the inputs consist of the LSTM hidden state of the model, and the outputs are probabilities of edge connections, as well as the mean and variance components $\mu_k$ and $\sigma^2_k$ introduced in \eqref{eq:spnorm}. For notational simplicity, let $\mathbf{h}_G$ denote the current hidden state $\mathbf{h}$ summarizing the graph generated thus far. To compute edge existence probabilities, we set $p = \sigma(f_{p}(\mathbf{h}_G))$, where $f_{p}$ is the MLP that outputs a scalar that is mapped to a probability by the sigmoid function $\sigma$. Similarly, the softplus-transformed normal parameters $\mu_{k}$ and $\sigma^2_{k}$ are parameterized with MLPs $f_\mu$ and $f_{\sigma^2}$, respectively: $\mu_{k} = f_\mu(\mathbf{h}_G)$ and $\sigma^2_{k} = \exp (f_{\sigma^2}(\mathbf{h}_G))$.

\paragraph{Tree-LSTM Cells} Finally, we note that BiGG offers a scalable approach to the typically slow process of training on graph data by leveraging an algorithm with $\mathcal{O}(\log n)$ runtime. This efficiency is achieved with the usage of a Tree-LSTM cell \citep{tai}, a variant of the standard LSTM cell that is designed for tree-structured inputs. Specifically, BiGG and its extensions employ a binary Tree-LSTM cell, where each internal node has exactly two children. Throughout this work, we refer to these simply as Tree-LSTM cells, with the understanding that they are all binary in structure.

Each node in the tree is associated with a hidden state $(\mathbf{h}_{j}, \mathbf{c}_{j})$, where $j$ denotes the index of the corresponding node. Leaf nodes consist of states computed directly from the model, of which no computations are made within the Tree-LSTM cell. For internal nodes, which have two children designated as the left child ($L$) and right child ($R$), the summary state at node $j$ is computed using the states of its children, $jL$ and $jR$, as follows \citep{tai}:

\begin{itemize}[leftmargin=1in]
    \setlength\itemsep{-0.0em}
    \item $\mathbf{i}_j = \sigma(\mathbf{W}_{L}^{(i)}\mathbf{h}_{jL} + \mathbf{W}_{R}^{(i)}\mathbf{h}_{jR} + \mathbf{b}^{(i)})$
    \item $\mathbf{f}_{jL} = \sigma(\mathbf{W}_{L}^{(fL)}\mathbf{h}_{jL} + \mathbf{W}_{R}^{(fL)}\mathbf{h}_{jR} + \mathbf{b}^{(f)})$
    \item $\mathbf{f}_{jR} = \sigma(\mathbf{W}_{L}^{(fR)}\mathbf{h}_{jL} + \mathbf{W}_{R}^{(fR)}\mathbf{h}_{jR} + \mathbf{b}^{(f)})$
    \item $\mathbf{g}_j = \tanh(\mathbf{W}_{L}^{(g)}\mathbf{h}_{jL} + \mathbf{W}_{R}^{(g)}\mathbf{h}_{jR} + \mathbf{b}^{(g)})$
    \item $\mathbf{o}_j = \sigma(\mathbf{W}_{L}^{(o)}\mathbf{h}_{jL} + \mathbf{W}_{R}^{(o)}\mathbf{h}_{jR} + \mathbf{b}^{(o)})$
    \item $\mathbf{c}_j = \mathbf{i}_j \odot \mathbf{g}_j + \mathbf{f}_{jL} \odot \mathbf{c}_{jL} + \mathbf{f}_{jR} \odot \mathbf{c}_{jR}$
    \item $\mathbf{h}_j = \mathbf{o}_j \odot \tanh(\mathbf{c}_j)$
\end{itemize}
A few key differences are worth noting. Unlike the standard sequential LSTM, each node in the tree maintains its own hidden state but does not receive an input vector $\mathbf{x}$. Additionally, the Tree-LSTM cell computes distinct forget gates for the left and right children --- denoted $\mathbf{f}_{jL}$ and $\mathbf{f}_{jR}$, respectively --- which allows the model to  attend to different information from each child node \citep{tai}. We use the notation $\mathbf{h}_{j} = \textrm{TreeLSTM}(\mathbf{h}_{jL}, \mathbf{h}_{jR})$ to represent the computation at internal node $j$ based on its two children, where the use of the corresponding cell states is implied. In this way, the Tree-LSTM cell represents a mechanism for merging two input states into a single summary state that captures pertinent information from both children.

Finally, we note that all LSTM cells, MLPs, and Tree-LSTM cells are parameterized by neural network parameters $\boldsymbol{\theta}$, which are optimized using the objective functions given in \eqref{eq:adjobj} (Adj-LSTM) and \eqref{eq:biggobj} (BiGG models) via gradient descent when training each model on a collection of training graphs.

\subsubsection{\texorpdfstring{Further Details on BiGG \citep{dai}}{Further Details on BiGG}} \label{sec:biggexpand}

%\subsection{Further Details on BiGG \citep{dai}} \label{sec:biggexpand}

BiGG generates an unweighted graph with an algorithm decomposed into two main components, both of which train in $\mathcal{O}(\log n)$ time: (1) row generation, where BiGG generates each row of the lower half of $\mathbf{A}$ using a binary decision tree; and (2) row conditioning, where BiGG deploys a hierarchical data maintenance structure called a Fenwick Tree generate each row conditioned on all prior rows.

\subsubsection{Row Generation} \label{sec:rowgenapp}

For each node $v_u \in G$, BiGG constructs a binary decision tree $\mathcal{T}_{u}$, adapted from R-MAT \citep{chakrabarti}, to determine $v_u$'s edge connections with the previously sequenced nodes $v_1, \ldots, v_{u-1}$. Rather than evaluating each potential connection sequentially, which incurs a cost of $\mathcal{O}(n)$ per row of $\mathbf{A}$, BiGG first checks whether any edge exists within the interval $[v_1, v_{u-1}]$. If so, it applies the R-MAT procedure to recursively divide this interval in half and identify the specific nodes connected to $v_u$. By partitioning the interval in halves rather than scanning all possible connections, BiGG reduces the row generation time to $\mathcal{O}(\log n)$. 

Let $t \in \mathcal{T}_{u}$ correspond to the node interval $[v_j, v_k]$ of length $l_t=k-j+1$, and denote the left-half and right-half of this interval as lch($t) = [v_j, v_{j+\floor{l_{t}/2}}]$ and rch($t) = [v_{j+\floor{l_{t}/2}+1}, v_k]$, respectively. To recursively generate $\mathcal{T}_{u}$, BiGG uses the following procedure for each $t \in \mathcal{T}_{u}$. If $t$ is a leaf in the tree --- that is, $t$ corresponds to a trivial interval of a single node $v_j$ --- an edge forms between $v_j$ and $v_u$. Otherwise, BiGG considers whether an edge exists in lch($t$). If so, the model recurses into lch($t$) until reaching a leaf. After constructing the entire left subtree, the model considers whether an edge exists in $\textrm{rch}(t)$, conditioned on this left subtree. If so, the model recurses into  rch($t$) until reaching a leaf. All leaves in the fully constructed tree thus represent nodes which connect with $v_u$ in the graph.

Each decision about the presence of an edge in the interval corresponding to node $t$ is modeled as a Bernoulli random variable with probability as a function of $\boldsymbol{\theta}$. To make these predictions autoregressive, BiGG maintains two context vectors. The top-down context $\mathbf{h}^{\text{top}}_u(t)$ encodes the history of all previous left-right decisions along the path from the root to the node $t$ and is used to estimate the probability of an edge in the left child, $\textrm{lch}(t)$. The bottom-up context $\mathbf{h}^{\text{bot}}_u(t)$ captures information from the already generated left subtree rooted at node $t$, and is used to condition the probability of an edge in the right child, $\textrm{rch}(t)$, based on the structure of $\textrm{lch}(t)$. Since $\textrm{rch}(t)$ is constructed only after $\textrm{lch}(t)$ and its dependencies, BiGG employs a Tree-LSTM cell \citep{tai} at node $t$ to merge the top-down context with the bottom-up summary of the left subtree prior to generating rch($t$). Hence, the probability of constructing each tree $\mathcal{T}_{u}$ is

\begin{equation} \label{eq:tuprob}
p_{\boldsymbol{\theta}}(\mathcal{T}_u ) =  \prod_{t\in \mathcal{T}_u}p_{\boldsymbol{\theta}}(\textrm{lch}(t) | \mathbf{h}^{top}_u(t)) \cdot  p_{\boldsymbol{\theta}}(\textrm{rch(t)} | \mathbf{\hat{h}}^{top}_u(t)),
\end{equation}

where $\mathbf{\hat{h}}^{top}_u(t) = \textrm{TreeCell}_{\boldsymbol{\theta}}(\mathbf{h}^{\text{top}}_u(t), \mathbf{h}^{\text{bot}}_u(\textrm{lch}(t)))$, and conditioning $\textrm{lch}(t)$ and $\textrm{rch}(t)$ on the top-down and bottom-up context vectors allows BiGG to generate all of $\mathcal{T}_{u}$ autoregressively. Equation~\ref{eq:tuprob} also implies that the probabilities of edges $p_{\boldsymbol{\theta}}(e_k)$ in Equation~\ref{biggedge} are modeled as a sequence of left-right edge existence Bernoulli decisions, which are then included in the objective function for all BiGG models. Figure~\ref{fig:topfig} provides an example of constructing $\mathcal{T}_u$ and visualizes the top-down and bottom-up states used to predict left and right edge existence decisions.

\subsubsection{Fenwick Tree} \label{sec:fent} The construction of $\mathcal{T}_{u}$ in Section~\ref{sec:rowgenapp} enables modeling the dependencies among edge connections within a single row of $\mathbf{A}$. To make BiGG fully autoregressive, however, it must also capture dependencies \textit{between} rows. For this purpose, BiGG incorporates the Fenwick tree \citep{fenwick} — a data structure designed to efficiently maintain prefix sums by performing sum operations over an array of length $n$ in $\mathcal{O}(\log n)$ time. To maintain a history of all previously generated rows in the graph, BiGG modifies the Fenwick tree to store summary representations of each row generated so far. This structure enables BiGG to compute a summary state of the first $u$ rows in $\mathcal{O}(\log u)$ time. 

The Fenwick topology tree is structured into $\lfloor \log(u - 1) \rfloor + 1$ levels. At the base—level 0—the leaves of the tree represent independent row embeddings for each $\mathcal{T}_{u}$. These embeddings are constructed using a bottom-up traversal: information starting from the leaf nodes of $\mathcal{T}_{u}$ propagates upward through the tree, culminating in a root-level summary that captures the overall structure of $\mathcal{T}_{u}$ and therefore summarizes all node connections with $v_u$. Entries along higher levels of the Fenwick topology tree represent merged state summaries — each combining information from multiple lower-level embeddings to produce a collective summary state across several rows.

Let $\mathbf{g}^{i}_j = (\mathbf{h}, \mathbf{c})$ denote the hidden and cell states represented by $j$-th node on the $i$-th level $\mathbf{g}^{i}_j$ of the Fenwick topology tree. Any given non-leaf node $\mathbf{g}^{i}_j$ consists of merging two children nodes one level below as

\begin{equation} \label{fen1}
\mathbf{g}_{j}^{i} = \textrm{TreeCell}_{\boldsymbol{\theta}}^{row}(\mathbf{g}^{i-1}_{2j-1}, \mathbf{g}^{i-1}_{2j}).
\end{equation}
Here, $0 \le i \le \lfloor \log(u - 1) \rfloor + 1$ and $1 \le j \le \lfloor \frac{u}{2^i} \rfloor$, where $\mathbf{g}_{j}^{0}$ denotes the bottom-up summary state of $\mathcal{T}_j$. To compute a representation of all previously generated rows at each iteration, the model iteratively applies another Tree-LSTM cell over the relevant summaries to produce the row-level summary state $\mathbf{h}^{\text{row}}_u$:

\begin{equation} \label{fen2}
\mathbf{h}^{\text{row}}_u = \textrm{TreeCell}_{\boldsymbol{\theta}}^{\textrm{summary}}\bigg(\bigg[\mathbf{g}^{i}_{\floor{\frac{k}{2^{i}}}}\ \textrm{where}\ k\ \&\ 2^{i} = 2^{i} \bigg]\bigg)
\end{equation}
where \& is the bit-level `and' operator. Thus, the Fenwick topology tree enables BiGG to generate rows in an autoregressive manner: the hidden state $\mathbf{h}_u^{\text{row}}$, defined in equation \eqref{fen2}, contains a history of all rows generated in $\mathbf{A}$ so far and is used as the initial state for constructing the next decision tree, $\mathcal{T}_{u+1}$. Similarly, during training, the binary structure of the Fenwick row tree allows the initialization of row states for each row of $\mathbf{A}$ to be computed in $\mathcal{O}(\log n)$ time.

\subsubsection{BiGG Training Procedure} \label{sec:bigg-train-app}

BiGG divides the training procedure into four main components, which all run on $\mathcal{O}(\log n)$ time:

\begin{enumerate}
    \item First, because all decision trees $\mathcal{T}_u$ are known a priori during training, bottom-up summaries are computed from the leaves to the root of each tree. Importantly, the summary state for each row is independent of those for other rows, allowing these computations to be performed in parallel.
    \item Next, using the root-level summaries from each $\mathcal{T}_u$, the model computes internal nodes $\mathbf{g}_j^{i}$ of the Fenwick topology tree in a level-wise manner using \eqref{fen1}.
    \item Once the Fenwick topology tree is constructed, BiGG concurrently computes the row summary states $\mathbf{h}_{u}^{row}$ in parallel using \eqref{fen2}.
    \item Finally, given each row-level summary $\mathbf{h}_{u}^{\text{row}}$, the model computes all left and right edge existence probabilities in parallel, traversing from the root to the leaves of each tree, as defined in \eqref{eq:tuprob}.
\end{enumerate}
Finally, we note that for graph generation, the trees $\mathcal{T}_{u}$ must be constructed sequentially. However, each $\mathcal{T}_{u}$ can still be built in $\mathcal{O}(\log n)$ time, reducing the overall runtime from $\mathcal{O}(n^2)$ to $\mathcal{O}((n + m)\log n)$.

\subsection{Adjacency-LSTM Motivation} \label{sec:adjlstm}

The primary issue with using an LSTM to build the adjacency matrix of a graph is that most recurrent neural networks are best suited for linear data. Flattening an adjacency matrix and using an LSTM is one potential avenue for building a model, but suffers from significant drawbacks -- the flattened vector varies based upon the node ordering $\pi$, and the model sacrifices the underlying structure of $\mathbf{A}$. \citep{you} Generative models such as GraphRNN and GRAN, which use recurrent neural networks to build generative models of graphs, circumvent this issue by using node-level and graph-level recurrent networks that maintain edge generation and the global structure of the graph, respectively. Adj-LSTM was inspired by such methods and instead uses a partitioning of the hidden state to take advantage of the grid structure of the adjacency matrix directly. We will also show that partitioning the hidden state of a single LSTM provides greater generative quality, as this facilitates information passing between the states of the row and column nodes.

\begin{algorithm}[t]
    \caption{Adjacency-LSTM Sampling Algorithm}
  \begin{algorithmic}[1]
    \INPUT Number of nodes $n$
    \STATE \textbf{Initialization} Initial row node state $\mathbf{s}^{R}_{0,0} = (\mathbf{h}_{0,0}, \mathbf{c}_{0,0})$
    \FOR{$i = 1,...,n$}
      \STATE $s_{i0} = s_{i-1,i-1} + \textrm{Pos}(n+1 - i)$
        \COMMENT{initialize new row node state}
      \FOR{$j < i$}
      \STATE $\mathbf{s}_{ij} = \textrm{Cat}(\mathbf{s}_{i, j - 1}^{R}, \mathbf{s}_{i - 1, j}^{C})$
        \COMMENT{concatenate previous node states of row $i$ and column $j$}
      \STATE $p_{ij} = \sigma(f_{p}(\mathbf{h}_{ij}))$
      \STATE Sample edge $e_{ij} \sim \textrm{Bernoulli}(p_{ij})$
      \IF{edge exists}
       \STATE $\mu_{ij} = f_{\mu}(\mathbf{h}_{ij})$
       \STATE $\log \sigma^{2}_{ij} = f_{\sigma^2}(\mathbf{h}_{ij})$
       \STATE $\epsilon_{ij} \sim \textrm{Normal}(\mu_{ij}, \sigma_{ij})$
       \STATE $w_{ij} = \log(1+\exp(\epsilon_{ij}))$
      \ELSE
       \STATE $w_{ij} = 0$
      \ENDIF
      \STATE $\textrm{embed}(e_{ij}, w_{ij}) = \textrm{Cat}(E_{ij}, f_{w}(w_{ij}), \textrm{Pos}(n + 1 -i), \textrm{Pos}(n - 1 + j))$
      \STATE $\mathbf{s}_{ij}^{*} = \textrm{LSTM}(\textrm{embed}(e_{ij}, w_{ij}); \mathbf{s}_{ij})$
        \COMMENT{update adjacency state with edge embedding}
      \STATE $\mathbf{s}_{ij}^{R}, \mathbf{s}_{ij}^{C} = \textrm{Split}(\mathbf{s}_{ij}^{*})$
      \ENDFOR
      \STATE $\mathbf{s}^{R}_{i,i} = \mathbf{s}^{R}_{i,i-1}$
        \COMMENT{set final row node state for subsequent row generation}
    \ENDFOR
    \OUTPUT $G$ with $V = \{1, 2, ... , n\}$ and $E = \{e_{ij}, w_{ij}\}_{i = 1; j > i}^{n}$
  \end{algorithmic}
\end{algorithm}

To adapt the LSTM architecture to a two-dimensional adjacency matrix, we partition the hidden state of the LSTM into $\mathbf{h}_{ij} = \begin{bmatrix} \mathbf{h}_{i,j-1}^{R} \\ \mathbf{h}_{i-1,j}^{C} \end{bmatrix}$, where $\mathbf{h}_{i,j-1}^{R}$ and $\mathbf{h}_{i-1,j}^{C}$ are the prior hidden states of the row and column nodes corresponding to entry $A_{ij}$. As all state updates are the same regardless of which entry $A_{ij}$ is being generated, we drop the subscripts $i$ and $j$ moving forward.

We re-compute the linear recurrence (4) using this partitioning of the hidden state. First, we note the dimensions of each weight and bias vector. Suppose that the hidden dimension is $h_{dim}$ and the embedding dimension is $i_{dim}$. Then we have the following:

\begin{enumerate}
 \item $\mathbf{h}^{R}, \mathbf{h}^{C} \in \mathbb{R}^{h_{dim}} \implies \mathbf{h} \in \mathbb{R}^{2h_{dim}}$ and $\mathbf{W}_{h} \in \mathbb{R}^{2h_{dim} \times 2h_{dim}}$. 
 \item $\mathbf{x}_{ij} \in \mathbb{R}^{i_{dim}} \implies \mathbf{W}_{i} \in \mathbb{R}^{2h_{dim} \times i_{dim}}$. \item $\mathbf{b} \in \mathbb{R}^{2h_{dim}}$. 
\end{enumerate}

Hence, we can partition the weight matrices and bias vector from the LSTM equations by defining the following partitions for each weight matrix:

\begin{equation} \label{eq:xmatp}
\mathbf{W}_{x} = \begin{bmatrix}
           \mathbf{U}^{R} \\
           \mathbf{U}^{C}
         \end{bmatrix}
\end{equation}

\begin{equation} \label{eq:hmatp}
\mathbf{W}_{h} = \begin{bmatrix}
           \mathbf{V}^{RR} & \mathbf{V}^{RC} \\
           \mathbf{V}^{CR} & \mathbf{V}^{CC}
         \end{bmatrix}
\end{equation}

where each $\mathbf{U}^{*} \in \mathbb{R}^{h_{dim} \times i_{dim}}$ and each $\mathbf{V}^{**} \in \mathbb{R}^{h_{dim} \times h_{dim}}$.

Using the partitioning from \ref{eq:xmatp} and \ref{eq:hmatp} and partitioning the bias vector as $\mathbf{b} = \begin{bmatrix} \mathbf{b}^{R} \\ \mathbf{b}^{C} \end{bmatrix}$, we can partition the LSTM update equations as

\[
\begin{bmatrix} 
    \mathbf{\hat{h}}^{R} \\
    \mathbf{\hat{h}}^{C}
\end{bmatrix}
= 
\begin{bmatrix}
   \mathbf{U}^{R} \mathbf{x} + \mathbf{V}^{RR} \mathbf{h}^{R} + \mathbf{V}^{RC} \mathbf{h}^{C} + \mathbf{b}^{R} \\
   \mathbf{U}^{C}x + \mathbf{V}^{CR} \mathbf{h}^{R} + \mathbf{V}^{CC} \mathbf{h}^{C} + \mathbf{b}^{C}
  \end{bmatrix}.
\]

Jointly updating the row and column states allows for the transfer of information between the row and column nodes via the weight matrices $\mathbf{V}^{RC}$ and $\mathbf{V}^{CR}$, which we hypothesized would mitigate the issue of long-term memory -- as the model is predicting entries row-wise, information early in the row-generation process becomes lossy without the joint update property of the LSTM with a partitioned hidden state.

To test this hypothesis, we trained the lobster graphs on two models: one which uses the single LSTM-update on the concatenated row and column states, and the other which uses two  LSTMs that update the row and the column states independently, which corresponds to setting $\mathbf{V}^{CR} = \mathbf{V}^{RC} = \mathbf{0}$ in Equation~\ref{eq:hmatp}. As observed in Table \ref{tab:tab4}, the LSTM joint update provides superior results on all observed metrics. 

\begin{table*}[t]
  \centering
    \caption[Joint State Update Performance]{Performance on updating the states simultaneously (``Joint``) vs  separately (``Independent``)}
  \begin{tabular}{ccccccc}
    \specialrule{1.25pt}{1pt}{1pt} \\[-2mm] 
    Update Mode & Deg. & Clus. & Top Spec. & Wt. Spec. & MMDWt & Error\\
    \midrule
    Joint & $2.46e^{-4}$ & $0.0$ & $9.98e^{-4}$ & $8.51e^{-4}$ & $4.93e^{-3}$ & 0.065
    \\
    Independent & $3.27e^{-4}$ & $6.20e^{-5}$ & $2.98e^{-3}$ & $2.46e^{-3}$ & $6.40e^{-3}$ & 0.275
    \\
    \bottomrule
  \end{tabular}
  \label{tab:tab4}
\end{table*}

\subsection{Tree Weight Generation}

The hierarchical sampling scheme of the tree weights provided a means of testing for autoregressiveness in the models with respect to the edge weights.  There are two main quantities of interest: the global variance of weights pooled across all trees, $\textrm{Var}(w_{ij})$, and the variance of weights found in a single tree, $\textrm{Var}(w_{ij} | \mu_{k}))$. Note a few preliminaries that are easily derived from their respective distributions

\begin{enumerate}
    \item $\mu_{k} \sim \mathcal{U}(7, 13) \implies \mathbb{E}(\mu_k) = 10$ and $\textrm{Var}(\mu_k) = 3$
    \item $w_{ij} \sim \Gamma(\mu_{k}^{2}, \mu_{k}^{-1}) \implies \mathbb{E}(w_{ij}|\mu_{k}) = \mu_{k}$ and  $\textrm{Var}(w_{ij}|\mu_{k}) = 1$
\end{enumerate}

Importantly, the variance of the weights found in each tree is free of the parameter $\mu_k$. Next, an application of iterative expectation and variance yield the mean and variance of weights pooled from all trees as

\begin{enumerate}
    \item $\mathbbm{E}(w_{ij}) = \mathbb{E}_{\mu}[\mathbb{E}_{w}(w_{ij}|\mu_{k})] = \mathbb{E}_{\mu}(\mu_{k}) = 10$.
    \item $\textrm{Var}(w_{ij}) = \mathbb{E}_{\mu}[\textrm{Var}_{w}(w_{ij}|\mu_{k})] + \textrm{Var}_{\mu}[\mathbb{E}_{w}(w_{ij}|\mu_{k})]  
     = \mathbb{E}_{\mu}(\mathbbm{1}) + \textrm{Var}_{\mu}(\mu_{k}) = 1 + 3 = 4$
\end{enumerate}

Thus, to test for autoregressiveness in the models, we observe that weights pooled from all trees have variance $\textrm{Var}(w_{ij}) = 4$, whereas weights from a single tree have variance $\textrm{Var}(w_{ij}|\mu_{k}) = 1$.

\subsection{Further Training Details}\label{ap7.4}

\paragraph{Hyperparameters}

For Adj-LSTM, node states were parameterized with a hidden dimension of 128 and use a 2-layer LSTM. An embedding dimension of 32 was used to embed edge existence, and an embedding dimension of 16 was used to embed the weights. Positional encoding was used on the initialized row states.

For all models using BiGG, we use a hidden dimension of 256 with position encoding on the row states, as used in the original BiGG model. The weight state we use for BiGG-E has a hidden dimension of 16 for model runs and 32 for scalability runs. For BiGG-E and BiGG+GCN, bits compression of 256 was used on the topology state.

\paragraph{Training Procedure}
Both models were trained using the Adam Optimizer with weight decay and an initial learning rate of $1e^{-3}$. The learning rate is decreased to $1e^{-5}$ when training loss plateus. Separate plateus are used for the weight parameters and the topology parameters.
\begin{itemize}

\item For the lobster data set, we train BiGG-E and BiGG-MLP for 250 epochs and validate at the 100 and 200th epochs. We plateau weight at epoch 50 and topology at epoch 100. We train Adj-LSTM for 300 epochs and validate every 100 epochs. We decay the learning rate at the 25 and 100th epochs.

\item For the tree data set, we train BiGG-E and BiGG-MLP for 550 epochs and validate every 100 epoch. We plateu weight at epoch 150 and topology at epoch 200. We train Adj-LSTM for 100 epochs and validate every 25 epochs, where the learning rate is plateaued at epochs 25 and 50.

\item For the Erdős–Rényi data set, BiGG-E was trained for 500 epochs and validated every 250 epochs. We plateu weight at epoch 100 and topology at epoch 500. Due to slow training and poor convergence, the Adjacency-LSTM was only trained for 27 epochs.

\item For the 3D Point Cloud data set, BiGG-E was trained for 3000 epochs and validated every 1000 epochs. We plateu weight at epoch 500 and topology at epoch 1500.Adj-LSTM was reported out of memory for this dataset.

\item For the joint graphs, BiGG-E was trained for 1000 epochs. We plateu weight and topology at 500 epochs.
\end{itemize}

BiGG-MLP and BiGG+GCN follow the training protocol for BiGG-E. For the convolutional network, two convolutions were used and each component was trained jointly on the same objective function used on BiGG-E.

\paragraph{Baseline Models}

The Erdős–Rényi model baseline estimates were generated by first estimating the global probability of an edge existing between two nodes based on the training data, and then constructing Erdős–Rényi graphs with that probability of edge existence, as done in \citep{you}. Weights were sampled with replacement from all possible training weights in order to produce weighted graphs.

\paragraph{Regularization}

We use the following to regularize our models, noting there is a propensity for the models to overtrain on the edge weights and decrease generative quality of the graphs. We believe this is directly related to the issue of balancing two losses - the topological and weight losses - which is further compounded by the fact learning the topology of a graph is much more challenging than the edge weights due to super-exponentially growing configuration space of graphs with respect to the number of nodes $n$. We note while BiGG-E had a tendency to overtrain on edge weights when plateauing the loss, the effect was much more prominent with BiGG-MLP, and regularization attempts were less successful with BiGG-MLP than on BiGG-E. As such, we use the following regularization on all models:

\begin{itemize}
    \item We use weight decay on the Adam optimizer with decay $1e^{-4}$ on topology parameters and $1e^{-3}$ on weight parameters.
    \item Prior to inputting the sampled weight through the embedding LSTM, we standardize the weight as $\hat{w} = s^{-1}_w(w - \bar{w})$, where $\bar{w}$ and $s_w$ and the mean and standard deviation of all training weights, respectively. We note this is because the model is general purpose and must handle weights of varying magnitudes, where larger weights can potentially saturate the output of the embedding LSTM.
    \item The loss for weight is scaled down by a factor of 10 to balance the topology and weight losses. Upone plateuing both sets of parameters, the scale was increased to a factor of 100 for all graphs except the joint graphs.
    \item When both losses were plateued, the weight loss was updated every other epoch instead of every epoch to allow more fine-tuning of the graph topology without encouraging overtraining on the edge weights.
\end{itemize}
\end{document}

%% file: math_commands.tex
\usepackage{amsmath,amsfonts,bm}

\def\eqref#1{equation~\ref{#1}}
\def\floor#1{\lfloor #1 \rfloor}
\def\1{\bm{1}}

\DeclareMathAlphabet{\mathsfit}{\encodingdefault}{\sfdefault}{m}{sl}
\SetMathAlphabet{\mathsfit}{bold}{\encodingdefault}{\sfdefault}{bx}{n}

%% file: main.bbl
\begin{thebibliography}{30}
\providecommand{\natexlab}[1]{#1}
\providecommand{\url}[1]{\texttt{#1}}
\expandafter\ifx\csname urlstyle\endcsname\relax
  \providecommand{\doi}[1]{doi: #1}\else
  \providecommand{\doi}{doi: \begingroup \urlstyle{rm}\Url}\fi

\bibitem[Albert \& Barabasi(2002)Albert and Barabasi]{albert}
Reka Albert and Albert-Laszlo Barabasi.
\newblock Statistical mechanics of complex networks.
\newblock \emph{Reviews of Modern Physics}, 74\penalty0 (1):\penalty0 47, 2002.
\newblock URL \url{http://link.aps.org/abstract/RMP/v74/p47}.

\bibitem[Baele et~al.(2025)Baele, Ji, Hassler, McCrone, Shao, Zhang, Holbrook,
  Lemey, Drummond, Rambaut, and Suchard]{beast}
Guy Baele, Xiang Ji, Gabriel~W. Hassler, John~T. McCrone, Yucai Shao, Zhenyu
  Zhang, Andrew~J. Holbrook, Philippe Lemey, Alexei~J. Drummond, Andrew
  Rambaut, and Marc~A. Suchard.
\newblock Beast x for bayesian phylogenetic, phylogeographic and phylodynamic
  inference.
\newblock \emph{Nature Methods}, 2025.
\newblock \doi{10.1038/s41592-025-02751-x}.
\newblock URL \url{https://doi.org/10.1038/s41592-025-02751-x}.

\bibitem[Barjuan et~al.(2025)Barjuan, Zheng, and Serrano]{barjuan}
Laia Barjuan, Muhua Zheng, and M~{\'A}ngeles Serrano.
\newblock The multiscale self-similarity of the weighted human brain
  connectome.
\newblock \emph{PLOS Computational Biology}, 21\penalty0 (4):\penalty0 1--20,
  04 2025.
\newblock \doi{10.1371/journal.pcbi.1012848}.
\newblock URL \url{https://doi.org/10.1371/journal.pcbi.1012848}.

\bibitem[Bellingeri et~al.(2023)Bellingeri, Bevacqua, Sartori, Turchetto,
  Scotognella, Alfieri, Nguyen, Le, Nguyen, and Cassi]{bellingeri}
M.~Bellingeri, D.~Bevacqua, F.~Sartori, M.~Turchetto, F.~Scotognella,
  R.~Alfieri, N.~K.~K. Nguyen, T.~T. Le, Q.~Nguyen, and D.~Cassi.
\newblock Considering weights in real social networks: A review.
\newblock \emph{Frontiers in Physics}, Volume 11 - 2023, 2023.
\newblock ISSN 2296-424X.
\newblock \doi{10.3389/fphy.2023.1152243}.
\newblock URL
  \url{https://www.frontiersin.org/journals/physics/articles/10.3389/fphy.2023.1152243}.

\bibitem[Chakrabarti et~al.(2004)Chakrabarti, Zhan, and Faloutsos]{chakrabarti}
Deepayan Chakrabarti, Yiping Zhan, and Christos Faloutsos.
\newblock \emph{R-MAT: A Recursive Model for Graph Mining}, pp.\  442--446.
\newblock SIAM International Conference on Data Mining, 2004.
\newblock \doi{10.1137/1.9781611972740.43}.
\newblock URL \url{https://epubs.siam.org/doi/abs/10.1137/1.9781611972740.43}.

\bibitem[Dai et~al.(2020)Dai, Nazi, Li, Dai, and Schuurmans]{dai}
Hanjun Dai, Azade Nazi, Yujia Li, Bo~Dai, and Dale Schuurmans.
\newblock Scalable deep generative modeling for sparse graphs.
\newblock In \emph{Proceedings of the 37th International Conference on Machine
  Learning (ICML)}, volume 119 of \emph{Proceedings of Machine Learning
  Research}, pp.\  2302--2312. PMLR, 2020.
\newblock URL \url{http://proceedings.mlr.press/v119/dai20b.html}.

\bibitem[Erdős \& Rényi(1959)Erdős and Rényi]{erdos}
Paul Erdős and A.~Rényi.
\newblock On random graphs.
\newblock \emph{Publicationes Mathematicae (Debrecen)}, 6:\penalty0 290, 1959.
\newblock URL
  \url{/brokenurl#snap.stanford.edu/class/cs224w-readings/erdos60random.pdf}.

\bibitem[Fagiolo et~al.(2010)Fagiolo, Reyes, and Schiavo]{fagiolo}
Giorgio Fagiolo, Javier Reyes, and Stefano Schiavo.
\newblock The evolution of the world trade web: a weighted-network analysis.
\newblock \emph{Journal of Evolutionary Economics}, 20\penalty0 (4):\penalty0
  479--514, 2010.
\newblock \doi{10.1007/s00191-009-0160-x}.
\newblock URL \url{https://doi.org/10.1007/s00191-009-0160-x}.

\bibitem[Fenwick(1994)]{fenwick}
Peter~M. Fenwick.
\newblock A new data structure for cumulative frequency tables.
\newblock \emph{Softw. Pract. Exper.}, 24\penalty0 (3):\penalty0 327–336,
  March 1994.
\newblock ISSN 0038-0644.
\newblock \doi{10.1002/spe.4380240306}.
\newblock URL \url{https://doi.org/10.1002/spe.4380240306}.

\bibitem[Gretton et~al.(2012)Gretton, Borgwardt, Rasch, Sch\"{o}lkopf, and
  Smola]{gretton}
Arthur Gretton, Karsten~M. Borgwardt, Malte~J. Rasch, Bernhard Sch\"{o}lkopf,
  and Alexander Smola.
\newblock A kernel two-sample test.
\newblock \emph{J. Mach. Learn. Res.}, 13\penalty0 (1):\penalty0 723--773,
  March 2012.
\newblock ISSN 1532-4435.
\newblock URL \url{http://dl.acm.org/citation.cfm?id=2503308.2188410}.

\bibitem[Grover et~al.(2019)Grover, Zweig, and Ermon]{grover}
Aditya Grover, Aaron Zweig, and Stefano Ermon.
\newblock Graphite: Iterative generative modeling of graphs.
\newblock In Kamalika Chaudhuri and Ruslan Salakhutdinov (eds.),
  \emph{Proceedings of the 36th International Conference on Machine Learning},
  volume~97 of \emph{Proceedings of Machine Learning Research}, pp.\
  2434--2444. PMLR, 09--15 Jun 2019.
\newblock URL \url{https://proceedings.mlr.press/v97/grover19a.html}.

\bibitem[Hochreiter \& Schmidhuber(1997)Hochreiter and Schmidhuber]{hochreiter}
Sepp Hochreiter and J{\"u}rgen Schmidhuber.
\newblock Long short-term memory.
\newblock \emph{Neural Computation}, 9\penalty0 (8):\penalty0 1735--1780, 1997.
\newblock \doi{10.1162/neco.1997.9.8.1735}.

\bibitem[Ingraham et~al.(2019)Ingraham, Garg, Barzilay, and Jaakkola]{ingraham}
John Ingraham, Vikas Garg, Regina Barzilay, and Tommi Jaakkola.
\newblock Generative models for graph-based protein design.
\newblock In H.~Wallach, H.~Larochelle, A.~Beygelzimer, F.~d\textquotesingle
  Alch\'{e}-Buc, E.~Fox, and R.~Garnett (eds.), \emph{Advances in Neural
  Information Processing Systems}, volume~32. Curran Associates, Inc., 2019.
\newblock URL
  \url{https://proceedings.neurips.cc/paper_files/paper/2019/file/f3a4ff4839c56a5f460c88cce3666a2b-Paper.pdf}.

\bibitem[Jo et~al.(2022)Jo, Lee, and Hwang]{jo}
Jaehyeong Jo, Seul Lee, and Sung~Ju Hwang.
\newblock Score-based generative modeling of graphs via the system of
  stochastic differential equations, 2022.
\newblock URL \url{https://arxiv.org/abs/2202.02514}.

\bibitem[Kawai et~al.(2019)Kawai, Mukuta, and Harada]{kawai}
Wataru Kawai, Yusuke Mukuta, and Tatsuya Harada.
\newblock {GRAM:} scalable generative models for graphs with graph attention
  mechanism.
\newblock \emph{CoRR}, abs/1906.01861, 2019.
\newblock URL \url{http://arxiv.org/abs/1906.01861}.

\bibitem[Keretsu \& Sarmah(2016)Keretsu and Sarmah]{keretsu}
Seketoulie Keretsu and Rosy Sarmah.
\newblock Weighted edge based clustering to identify protein complexes in
  protein--protein interaction networks incorporating gene expression profile.
\newblock \emph{Computational Biology and Chemistry}, 65:\penalty0 69--79,
  2016.
\newblock ISSN 1476-9271.
\newblock \doi{https://doi.org/10.1016/j.compbiolchem.2016.10.001}.
\newblock URL
  \url{https://www.sciencedirect.com/science/article/pii/S1476927115301857}.

\bibitem[Kipf \& Welling(2016)Kipf and Welling]{kipf}
Thomas~N. Kipf and Max Welling.
\newblock {Variational Graph Auto-Encoders}.
\newblock In \emph{NIPS Workshop on Bayesian Deep Learning}, BDL 2016, 2016.
\newblock URL \url{http://bayesiandeeplearning.org/2016/papers/BDL_16.pdf}.

\bibitem[Li et~al.(2018)Li, Vinyals, Dyer, Pascanu, and Battaglia]{li}
Yujia Li, Oriol Vinyals, Chris Dyer, Razvan Pascanu, and Peter~W. Battaglia.
\newblock Learning deep generative models of graphs.
\newblock \emph{CoRR}, abs/1803.03324, 2018.
\newblock URL \url{http://arxiv.org/abs/1803.03324}.

\bibitem[Liao et~al.(2019)Liao, Li, Song, Wang, Hamilton, Duvenaud, Urtasun,
  and Zemel]{liao}
Renjie Liao, Yujia Li, Yang Song, Shenlong Wang, William~L. Hamilton, David
  Duvenaud, Raquel Urtasun, and Richard~S. Zemel.
\newblock Efficient graph generation with graph recurrent attention networks.
\newblock In Hanna~M. Wallach, Hugo Larochelle, Alina Beygelzimer, Florence
  d'Alché Buc, Emily~B. Fox, and Roman Garnett (eds.), \emph{NeurIPS}, pp.\
  4257--4267, 2019.
\newblock URL
  \url{http://dblp.uni-trier.de/db/conf/nips/nips2019.html#LiaoLSWHDUZ19}.

\bibitem[Neumann et~al.(2013)Neumann, Moreno, Antanas, Garnett, and
  Kersting]{neumann}
M.~Neumann, P.~Moreno, L.~Antanas, R.~Garnett, and K.~Kersting.
\newblock {Graph Kernels for Object Category Prediction in Task-Dependent Robot
  Grasping}.
\newblock In \emph{Proceedings of the Eleventh Workshop on Mining and Learning
  with Graphs (MLG{--}2013)}, Chicago, US, 2013.

\bibitem[Niu et~al.(2020)Niu, Song, Song, Zhao, Grover, and Ermon]{niu}
Chenhao Niu, Yang Song, Jiaming Song, Shengjia Zhao, Aditya Grover, and Stefano
  Ermon.
\newblock Permutation invariant graph generation via score-based generative
  modeling, 2020.
\newblock URL \url{https://arxiv.org/abs/2003.00638}.

\bibitem[Qin et~al.(2024)Qin, Vignac, and Frossard]{sparsediff}
Yiming Qin, Clement Vignac, and Pascal Frossard.
\newblock Sparse training of discrete diffusion models for graph generation,
  2024.
\newblock URL \url{https://arxiv.org/abs/2311.02142}.

\bibitem[Rodríguez \& Dunson(2011)Rodríguez and Dunson]{rodriguez}
A.~Rodríguez and DB. Dunson.
\newblock Nonparametric bayesian models through probit stick-breaking
  processes.
\newblock \emph{Bayesian Anal.}, 2011.
\newblock [Accessed 11-09-2024].

\bibitem[Rumelhart et~al.(1986)Rumelhart, Hinton, and Williams]{rumelhart}
David~E Rumelhart, Geoffrey~E Hinton, and Ronald~J Williams.
\newblock Learning representations by back-propagating errors.
\newblock \emph{Nature}, 323\penalty0 (6088):\penalty0 533--536, 1986.
\newblock \doi{10.1038/323533a0}.

\bibitem[Sak et~al.(2014)Sak, Senior, and Beaufays]{sak}
Haşim Sak, Andrew Senior, and Françoise Beaufays.
\newblock Long short-term memory based recurrent neural network architectures
  for large vocabulary speech recognition, 2014.
\newblock URL \url{https://arxiv.org/abs/1402.1128}.

\bibitem[Semple et~al.(2003)Semple, Steel, and Steel]{semple}
C.~Semple, M.~Steel, and B.D.M.S.M. Steel.
\newblock \emph{Phylogenetics}.
\newblock Oxford lecture series in mathematics and its applications. Oxford
  University Press, 2003.
\newblock ISBN 9780198509424.
\newblock URL \url{https://books.google.com/books?id=uR8i2qetjSAC}.

\bibitem[Stanojevic et~al.(2018)Stanojevic, Abbar, and Mokbel]{Stanojevic}
Rade Stanojevic, Sofiane Abbar, and Mohamed Mokbel.
\newblock W-edge: weighing the edges of the road network.
\newblock In \emph{Proceedings of the 26th ACM SIGSPATIAL International
  Conference on Advances in Geographic Information Systems}, SIGSPATIAL '18,
  pp.\  424–427, New York, NY, USA, 2018. Association for Computing
  Machinery.
\newblock ISBN 9781450358897.
\newblock \doi{10.1145/3274895.3274916}.
\newblock URL \url{https://doi.org/10.1145/3274895.3274916}.

\bibitem[Tai et~al.(2015)Tai, Socher, and Manning]{tai}
Kai~Sheng Tai, Richard Socher, and Christopher~D. Manning.
\newblock Improved semantic representations from tree-structured long
  short-term memory networks.
\newblock \emph{CoRR}, abs/1503.00075, 2015.
\newblock URL \url{http://arxiv.org/abs/1503.00075}.

\bibitem[Vignac et~al.(2023)Vignac, Krawczuk, Siraudin, Wang, Cevher, and
  Frossard]{vignac}
Clement Vignac, Igor Krawczuk, Antoine Siraudin, Bohan Wang, Volkan Cevher, and
  Pascal Frossard.
\newblock Digress: Discrete denoising diffusion for graph generation, 2023.
\newblock URL \url{https://arxiv.org/abs/2209.14734}.

\bibitem[You et~al.(2018)You, Ying, Ren, Hamilton, and Leskovec]{you}
Jiaxuan You, Rex Ying, Xiang Ren, William~L. Hamilton, and Jure Leskovec.
\newblock Graphrnn: A deep generative model for graphs.
\newblock \emph{CoRR}, abs/1802.08773, 2018.
\newblock URL
  \url{http://dblp.uni-trier.de/db/journals/corr/corr1802.html#abs-1802-08773}.

\end{thebibliography}
